\documentclass[10pt,twocolumn,letterpaper]{article}

\usepackage[pagenumbers]{cvpr} %

\usepackage[dvipsnames]{xcolor}

\definecolor{emb_color}{RGB}{252,224,225}
\definecolor{multi_head_attention_color}{RGB}{252,226,187}
\definecolor{add_norm_color}{RGB}{242,243,193}
\definecolor{ff_color}{RGB}{194,232,247}
\definecolor{softmax_color}{RGB}{203,231,207}
\definecolor{linear_color}{RGB}{220,223,240}
\definecolor{gray_bbox_color}{RGB}{243,243,244}

\newcommand{\coolname}{\textit{TULIP}}

\definecolor{cvprblue}{rgb}{0.21,0.49,0.74}

\usepackage[pagebackref,breaklinks,colorlinks,citecolor=cvprblue]{hyperref}

\usepackage{colortbl}
\usepackage{subcaption}
\usepackage{stfloats}
\usepackage{balance}
\usepackage{appendix}
\usepackage[accsupp]{axessibility} %

\newcommand{\PAR}[1]{\vskip2pt \noindent{\bf #1}}
\newcommand*{\FIGINSUPP}{def}  %
\def\paperID{10202} %
\def\confName{CVPR}
\def\confYear{2024}

\title{TULIP: Transformer for Upsampling of LiDAR Point Clouds}

\author{Bin Yang$^{1}$, Patrick Pfreundschuh$^{1}$, Roland Siegwart$^{1}$, Marco Hutter$^{2}$, Peyman Moghadam$^{3,4{\dagger}}$, Vaishakh Patil$^{2{\dagger}}$\\
$^{1}$Autonomous Systems Lab, ETH Zurich, Switzerland, $^{2}$Robotic Systems Lab, ETH Zurich, Switzerland \\
$^{3}$School of EER, Queensland University of Technology (QUT), Australia, $^{4}$Data61, CSIRO, Brisbane, Australia \\
{\tt\small \;\{{biyang,\;patripfr,\;rolandsi,\;mahutter,\;patilv}\}@ethz.ch, \;peyman.moghadam@csiro.au} 
}

\begin{document}
\maketitle

\def\thefootnote{$\dagger$}\footnotetext{Authors share last authorship. }
\begin{abstract}
LiDAR Upsampling is a challenging task for the perception systems of robots and autonomous vehicles, due to the sparse and irregular structure of large-scale scene contexts. Recent works propose to solve this problem by converting LiDAR data from 3D Euclidean space into an image super-resolution problem in 2D image space. Although their methods can generate high-resolution range images with fine-grained details, the resulting 3D point clouds often blur out details and predict invalid points. In this paper, we propose \coolname{}, a new method to reconstruct high-resolution LiDAR point clouds from low-resolution LiDAR input. We also follow a range image-based approach but specifically modify the patch and window geometries of a Swin-Transformer-based network to better fit the characteristics of range images. We conducted several experiments on three public real-world and simulated datasets. \coolname{} outperforms state-of-the-art methods in all relevant metrics and generates robust and more realistic point clouds than prior works. The code is available at \url{https://github.com/ethz-asl/TULIP.git}.
\end{abstract}

\section{Introduction}
\label{sec:intro}

\begin{figure}[htbp]
  \centering
  \begin{subfigure}[b]{0.23\textwidth}
    \includegraphics[width=\textwidth]{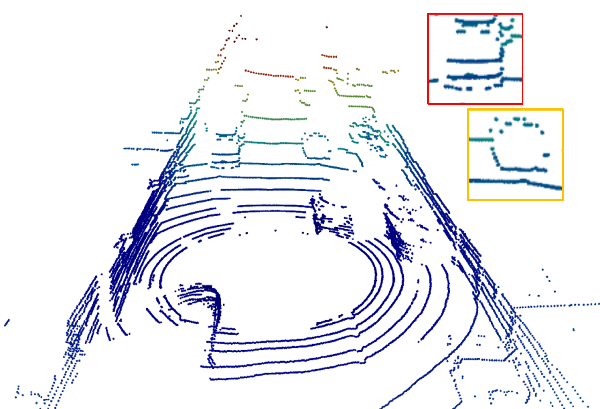}
    \caption{Input}
     \end{subfigure}
   \begin{subfigure}[b]{0.23\textwidth}
  \centering
    \includegraphics[width=\textwidth]{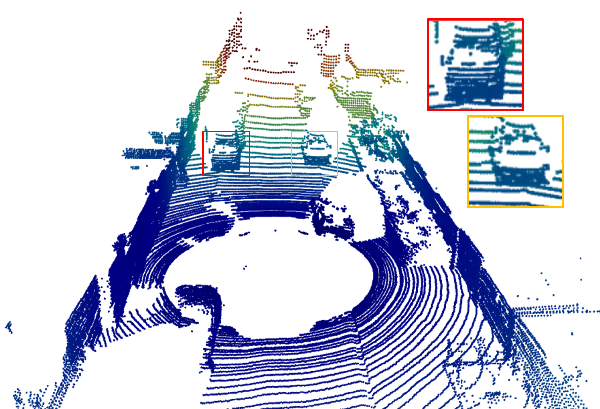}
    \caption{Ground Truth}
  \end{subfigure}
  \begin{subfigure}[b]{0.23\textwidth}
    \includegraphics[width=\textwidth]{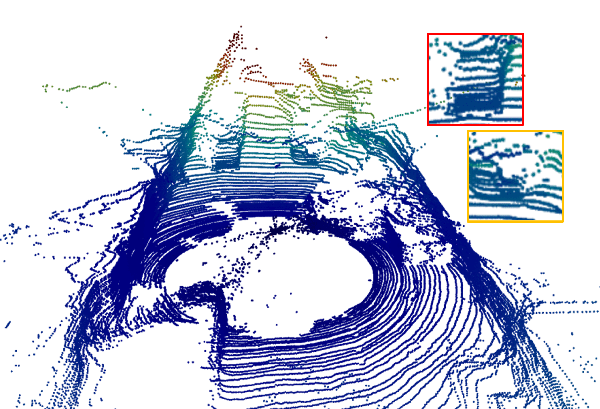}
    \caption{SwinIR~\cite{liang2021swinir}}
  \end{subfigure}
  \begin{subfigure}[b]{0.23\textwidth}
    \includegraphics[width=\textwidth]{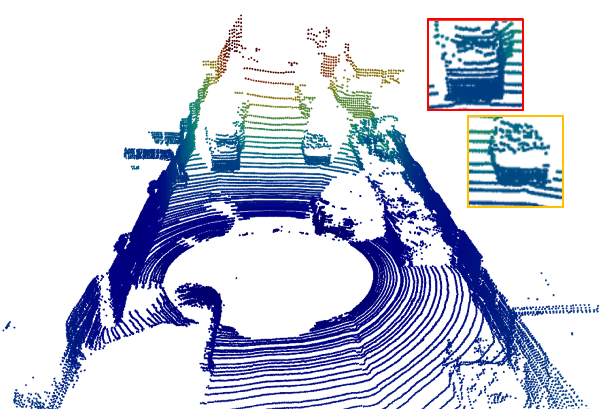}
    \caption{\coolname{} (Ours)}
  \end{subfigure}
  \vspace{-3mm}
  \caption{\coolname{} generates more realistic LiDAR point clouds from low-resolution LiDAR input (a) and outperforms the state-of-the-art image super-resolution approach~\cite{liang2021swinir} in upsampling with fine-grained details of different objects in the scene.}
  \label{fig:intro}
  \vspace{-6mm}
\end{figure}
Light Detection And Ranging (LiDAR) is one of the most common sensors for perception in various fields of autonomy, such as autonomous driving, and unmanned aerial vehicles (UAV). LiDARs are used to generate 3D point clouds of the scene. These point clouds are essential for mapping, localization, and object detection tasks. However, the accuracy of these tasks often depends on the resolution of the point cloud~\cite{yang2019std}. Furthermore, the resolution of a LiDAR is inherently associated with increased energy consumption and cost, making its use impractical for various applications. LiDARs also have different vertical and horizontal resolutions. The vertical resolution of rotating 3D LiDARs is typically much lower than the horizontal resolution. These limitations make an upsampling technique necessary to increase the resolution of the LiDAR data, especially in the vertical direction. Moreover, upsampling LiDAR data has the potential to counter domain shift problems in LiDAR-based learning methods. Notably, the use of lower resolution LiDAR data in a system that has been trained on higher resolution data leads to a significant drop in performance~\cite{bevsic2022unsupervised}. Therefore, such upsampling techniques can not only help to mitigate the domain gap by creating a virtual high-resolution sensor that matches the target domain~\cite{savkin2022lidar, zhao2021sspu}  but also reduce the high cost of collecting new LiDAR data, annotating and retraining the methods.\\
Deep learning has led to remarkable advancements in LiDAR upsampling techniques in recent years. Several recent methods~\cite{yu2018pu, zhang2019data, savkin2022lidar} have focused on learning the upsampling process within the 3D Euclidean space. However, the processing of 3D data in deep learning can be quite resource-intensive. One way to address this challenge is the representation of the point cloud as a range image~\cite{shan2020simulation, kwon2022implicit, eskandar2022hals, wang2023swintnfc}. This approach facilitates the adoption of well-established image-based super-resolution methods. Such methods mostly employ autoencoder-style networks with convolutional and deconvolutional layers~\cite{shan2020simulation, zovathi2023stdepthnet, dong2015imagesr}. However, these methods are not directly transferable to range images. Range images generated by the projection of sparse 3D point clouds have distinct sharp edges at the object boundaries. Convolutional neural network-based methods often induce edge smoothing due to the regularization effect, which limits their usability for LiDAR range image processing. \\
As an alternative, Transformer~\cite{vaswani2017attention} has shown great success in RGB image super-resolution and produces less blurry outputs~\cite{liang2021swinir}. Vision transformers, however, face challenges such as the need for large amounts of training data~\cite{chen2023hat, liang2021swinir, lu2022transformerimagesr} and the global computation of self-attention, limiting the capture of local information in the range images. To address this, Swin Transformer~\cite{liu2021swintransformer} introduced tokenizing the input image and applying the self-attention mechanism locally in independent windows. 
While the modifications to Swin Transformer have led to robust training and promising results compared to Vision Transformer, its default structure and settings are not ideally suited for processing LiDAR range images. These range images consist of single-channel data, representing 3D spatial information, as opposed to three-channel color images that depict visual appearance. Furthermore, range images feature large, smooth areas within 3D objects, separated by distinct sharp edges between objects. As a result, minor inaccuracies in the 2D prediction (\eg{} blurry output) can lead to significant differences in the projected 3D occupancy. This non-uniform distribution of image details further complicates network training. The challenging details in the range image are concentrated in a few specific pixels, contrasting with RGB images where relevant details are evenly distributed. This makes range images fundamentally different from RGB images.
Inspired by the limitations of state-of-the-art range image upsampling methods and the potential of Swin-Transformer~\cite{liu2021swintransformer}, we propose a novel network named \coolname{}. Our proposed geometry-aware architecture is tailored to accommodate LiDAR range image data. 
The proposed method takes a low-resolution range image as input and produces a high-resolution range image, which can then be projected into a 3D point cloud. To accommodate LiDAR range image geometry, we utilize one-line row patches to tokenize the input range image, as opposed to the square patches used for RGB images. 
This approach helps to preserve vertical information for the upsampling process while capturing boundary discontinuities effectively. This also creates a trade-off between local detail capture and model complexity in the training process. Additionally, we incorporate a non-square window for local self-attention computation. This enhances learning of spatial contexts at different scales in the range image. 
Further, we extensively train and evaluate \coolname{} on two real-world and one simulated autonomous driving datasets: KITTI~\cite{geiger2012KITTI}, DurLAR~\cite{li21durlar} and CARLA~\cite{kwon2022implicit}, respectively. 
Comparisons to existing approaches demonstrate that \coolname{} outperforms the state-of-the-art methods on all three datasets. %

\section{Related Works}
\label{sec:related_works}
\PAR{Image super-resolution} has seen significant progress in recent years. The technique aims to construct high-resolution (HR) images from low-resolution (LR) observations, often leveraging advances in Convolutional Neural Networks (CNNs) to enhance the fidelity and detail of images for better visualization and information extraction. Subsequently, several enhanced frameworks have been developed~\cite{dong2015imagesr, dong2016accelerating, zhang2020rdnir, kim2016deeply, kim2016accurate}. \\
Driven by their success in the field of natural language processing (NLP)~\cite{devlin2018bert, brown2020language}, Transformers~\cite{vaswani2017attention} have been extended to solve a variety of vision-related tasks such as object detection~\cite{carion2020detr} and semantic segmentation~\cite{amir2021deepvit, caron2021dino}. Such Vision Transformers (ViT)~\cite{dosovitskiy2020vit} excel at learning to focus on relevant image regions by exploring global interactions among different regions. Their impressive performance has led to their adoption in image restoration tasks~\cite{liang2021swinir, chen2023hat, chen2023dual} as well. While they brought significant improvements in RGB image super-resolution, Vision Transformer~\cite{dosovitskiy2020vit} comes with the drawback of the quadratic computational complexity of self-attention. Additionally, they capture mostly global dependencies within the data and require large amounts of data for training. \\
To address these challenges, Swin Transformer~\cite{liu2021swintransformer} has been proposed. Unlike the Vision Transformer, which relies on global self-attention across the entire image, the Swin Transformer employs a local window-based attention mechanism and establishes a pyramid-like architecture. This approach processes images hierarchically by gradually merging smaller image patches into larger ones. 
This strategy enables more effective handling of different scales and facilitates the processing of multi-scale features, making it particularly suitable for range image processing. Furthermore, due to Swin Transformer's hierarchical structure, the increase in the number of model parameters scales linearly with the input image size, which mitigates the challenges associated with handling large range images. Additionally, they require less training data than the classical Vision Transformer, which is beneficial for LiDAR upsampling as LiDAR datasets are typically magnitudes smaller than RGB datasets used for Vision Transformer.
The Swin and Vision Transformers have also been used for image super-resolution on omnidirectional panoramic camera images~\cite{sun2023opdn, yu2023osrt}. However, these approaches are specifically designed for the unique characteristics of these cameras (\eg{} strong distortions, spherical continuity) that are vastly different from the LiDAR sensing model, which renders them unsuitable for application to range images.
\PAR{Other related works} have focussed on upsampling high-density point clouds from low-density point clouds ~\cite{zhao2021sspu, yu2018pu, ye2021meta, zhao2022self, qiu2022pu}. While those approaches also provide qualitative results for LiDAR point clouds, they focus on increasing the general 3D point density of point clouds of single objects instead of a large scene. Another area closely related to LiDAR upsampling is depth completion~\cite{cheng2018depth, hu2021penet, yan2107rignet, tang2020learning, rho2022guideformer, qiu2019deeplidar, park2020non, lin2022dynamic}. The primary objective of depth completion is to improve sparse depth estimates acquired from LiDAR by integrating data from multiple sensors, predominantly RGB cameras. The result of depth completion methods is a dense depth map that provides depth values to every pixel in the input depth map. Nonetheless, this field slightly diverges from LiDAR upsampling due to its reliance on multimodal sensors and its common deployment within a restricted field of view, making it not directly comparable within the scope of this paper. 
LiDAR point clouds exhibit specific characteristics, such as a distinct point pattern (due to the stacked lasers in rotating 3D LiDARs) and a decrease in point density with increasing distance from the sensor. Differently from arbitrary point cloud upsampling, LiDAR upsampling tries to mimic a realistic point cloud of a high-resolution LiDAR given the point cloud of low-resolution LiDAR and thus targets a different result than the approaches above.
Due to the difficulty and high computational cost of upsampling LiDAR point clouds in 3D space~\cite{savkin2022lidar, akhtar2022pu, chen2022density}, most works ~\cite{kwon2022implicit, shan2020simulation, triess2019lidarcnn} represent the point cloud as a range image and perform the LiDAR upsampling task in 2D image space. LiDAR-CNN~\cite{triess2019lidarcnn} introduces a CNN-based architecture with semantic and perceptual guidance. The method specifies two further loss functions besides a per-point reconstruction loss to acquire a better synthesis of the high-resolution range image. However, this approach is limited to data with semantic annotations, which drastically reduces its applicability. LiDAR-SR~\cite{shan2020simulation} deploys a CNN-based network with U-Net~\cite{ronneberger2015u} architecture. It additionally uses Monte-Carlo dropout post-processing to reduce the amount of noisy points in the prediction. Approaches based on convolutional operations tend to fail to reconstruct the sharpness in the range image effectively. While the previous approaches directly predict the range image, Implicit LiDAR Network (ILN)~\cite{kwon2022implicit} uses an implicit neural architecture that learns interpolation weights to fill in new pixels instead of their values directly. Although the method outperforms CNN-based approaches in training speed and preserving the geometrical details in the input, it can still suffer from limited neighboring information in the LiDAR data, especially in areas distant from the sensor due to the extremely sparse input data. Differently from the previous works, our approach builds on Swin Transformer as a backbone and is specifically customized to effectively process range images.

\section{Methodology}
\label{sec:method}
LiDAR upsampling is achieved by upsampling a range image, effectively transforming it from a 3D upsampling problem into a 2D image super-resolution problem~\cite{kwon2022implicit, triess2019lidarcnn, shan2020simulation, eskandar2022hals}.
It is, therefore, an evident approach to build on RGB image super-resolution. However, image super-resolution aims to solve a different problem than LiDAR upsampling. Image super-resolution enhances the visual appearance of a low-resolution RGB image by recreating a high-resolution image. While quantitative metrics for evaluating these works exist, there is no one specific correct solution.
Differently, LiDAR upsampling tries to recreate a high-resolution, 3D LiDAR sensor output from a low-resolution LiDAR input point cloud. Range images contain more geometrical and spatial contexts rather than visual information, and the scene's geometry strictly dictates the correct solution. The underlying sensing model is also drastically different, as LiDARs are active sensors. The respective range pixel values purely depend on distance, while pixels in an RGB camera depend on many factors, such as scene appearance, lighting conditions, exposure time, and white balance.
Furthermore, range images have a highly asymmetric aspect ratio, typically between 1:8 to 1:64, which differs drastically from regular camera images.
Accordingly, most works~\cite{triess2019lidarcnn, eskandar2022hals, shan2020simulation} only perform vertical upsampling, in contrast to image super-resolution, which operates in both dimensions.
A naive application of state-of-the-art image super-resolution networks~\cite{liang2021swinir} to range images does not result in adequate performance (Fig.~\ref{fig:intro}).
To this end, we developed \coolname{}, a novel range image-based LiDAR Upsampling method. 
Based on the observations above, we specifically adapt a network based on SwinUnet~\cite{cao2022swinunet} to incorporate range images better. In the following sections, we will describe the technical details of our method, with a special focus on how our approach accommodates range images.

\subsection{Problem Definition}
A LiDAR point cloud consists of points captured during one revolution $\mathcal{P} = \{\mathbf{p}_1,...,\mathbf{p}_n\}$. Each measurement represent a 3D point $\mathbf{p}_i = \{x_i, y_i, z_i\}$. As input, we assume a low-resolution point cloud $\mathcal{P}_l$ with $n_l = H_l \times W_l$ points, where $H_l$ and $W_l$ correspond to vertical and horizontal resolution. We aim to predict a high-resolution point cloud $\mathcal{P}_h$, that has the same field-of-view(FoV) as $\mathcal{P}_l$ but contains $n_h = \beta * n_l$ points. In our experiments, we set $\beta = 4$. We only increase vertical resolution (amount of LiDAR beams), \ie{}, $n_h = H_h \times W_h$, where $H_h = \beta * H_l$ and $W_h=W_l$. 
We project the point cloud into a 2D range image. In this image, each row and column coordinate $v,u$ correspond to the respective elevation and azimuth angles of the LiDAR points, while the pixel value contains the range of the point $r = \sqrt{x^2 + y^2 + z^2}$. 
The image coordinate of a 3D point can be calculated using a spherical projection model:
\begin{equation} \label{eqn:prediction}
\begin{bmatrix} u \\ v \end{bmatrix} = 
\begin{bmatrix} \frac{W}{2}  - \frac{W}{2\pi} arctan(\frac{y}{x}) \\
 \frac{H}{\Theta_{max}-\Theta_{min}} * (\Theta_{max} - arctan(\frac{z}{\sqrt{x^2 + y^2}})
\end{bmatrix}
\end{equation}
$\Theta_{max}$ and $\Theta_{min}$ correspond to the limits of the vertical field of view of the LiDAR. We formulate the LiDAR upsampling problem as an image upsampling problem, \ie{} we aim to predict a high-resolution range image $I_{h} \in \mathbf{R}^{1\times H_h \times W}$ given a low-resolution range image $I_{l} \in \mathbf{R}^{1\times H_l \times W}$.
We can then calculate the high-resolution point cloud $\mathcal{P}_h$ by inverting the projection in Eq.~\ref{eqn:prediction}.

\subsection{Architecture Design and Overview}
Our network builds upon Swin-Unet~\cite{cao2022swinunet} which was originally designed for image segmentation. We deploy a U-shaped network structure featuring skip connections that link the encoder and decoder modules. The fundamental building block of \coolname{} is the Swin Transformer~\cite{liu2021swintransformer}. \coolname{} first patchifies and maps the input to a high-dimensional feature space. The tokenized low-level feature maps are passed through a multi-stage encoder. Each stage contains a pair of Swin Transformer~\cite{liu2021swintransformer} blocks and a patch merging layer to downsample the resolution by a factor of two and increase the dimension by a factor of four. This step realizes the hierarchic computation and extraction of multi-scale features through encoding. Within each stage, multiple instances of the self-attention mechanism are performed locally in parallel, which is the so-called Window Multi-Head Self Attention (W-MSA). After passing through a two-layer MLP and residual connection with the input, the feature vector is further passed to the second part of the Swin Transformer block, utilizing a Shifted Window Multi-Head Self Attention (SW-MSA) which extends W-MSA by employing a shifted window operation. This enhances the model capacity as it compensates for the lack of interaction between the local windows in W-MSA.
The decoder, which realizes the upsampling has a symmetric design to the encoder and operates in a reverse way of the encoder. The resolution of the feature maps is first expanded, and the dimension is reduced by a factor of two accordingly. We feed existing geometrical information via skip connections. 
The feature maps are transformed into a single-channel, high-resolution range image in the last layer. The reconstruction head comprises a $1\times1$ convolutional layer for feature expansion, followed by a Leaky ReLU activation, a pixel shuffle layer~\cite{shi2016pixelshuffle}, and another $1\times1$ convolutional layer for final projection. As the loss function, we select the pixel-wise L1 loss. Details of the network can be found in the Appendix.
\begin{figure*}[t!] %
    \centering
    \includegraphics[width=0.85\textwidth]{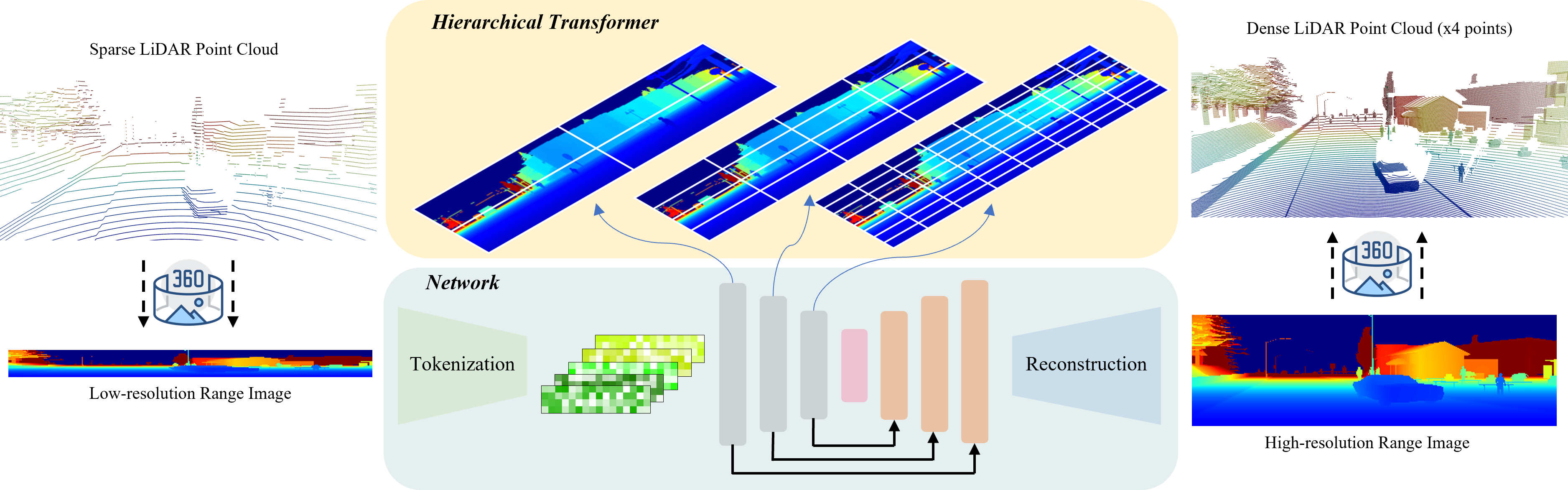}
    \caption{Overview of \coolname{}: The low-resolution point cloud is spherically projected into the range image. In the training and inference, the range image is tokenized into feature embeddings without compressing the spatial resolution along the vertical axis. We adopt the U-Net-based network presented in \cite{cao2022swinunet} for range image upsampling. We follow the hierarchical structure of Swin Transformer~\cite{liu2021swintransformer} for feature extraction and use the non-square window geometry for computing the local self-attention. The reconstruction head generates the high-resolution range image from the latent features and then, the point cloud can be obtained by projecting pixels back into the 3D space.}
    \vspace{-4mm}
    \label{fig:overview}
\end{figure*}

\subsubsection{Tokenization}
\label{subsec:patch partition}
Tokenization is done by a patch partition layer that creates an initial feature embedding from the input range image. This initial feature representation significantly influences the network's performance. Specifically, the selection of this layer determines how the inherent characteristics, patterns, and relevant information of the LiDAR range image data are encoded into a format that the network can effectively learn from. 
Most transformer-based RGB image super-resolution approaches utilize a relatively large patch size to tokenize the input~\cite{liang2021swinir, chen2023hat, lu2022transformerimagesr}. These larger patches enable the construction of a global spatial context, facilitating an understanding of the overall structure and high-level features of the RGB image. These networks employ square-shaped patches and implement upsampling in both vertical and horizontal directions.
In contrast, LiDAR upsampling primarily aims to enhance the vertical resolution of the input, given that range images possess properties that are vastly different from RGB images. 
Furthermore, attention across more pixels in range images is less useful than in RGB images, since there is almost no geometrical relation for the spatial contexts that are far from each other in the 2D range image space.
Motivated by this, we propose two adjustments to effectively process range images within a Swin-Unet: 

\PAR{Row-Based Patch Partition:} Our model builds row patches with a dimension of $1\times4$ for range image tokenization. The new patch geometry is designed to retain full vertical information while compressing horizontally. This aligns with our objective of conserving and extending vertical details in the range image. Besides that, row patches excel at capturing boundary discontinuities between distinct objects and the background within the scene. 
Moreover, the selection of the patch size strikes a balance between enhancing model performance and limiting model capacity. Smaller patches can capture more detailed local information, but they also increase the number of model parameters, which can slow down the model's training process.
\PAR{Circular Padding (CP) in Horizontal Dimension:} Instead of adding zero pixels padding, our model uses circular padding in the horizontal dimension. This avoids introducing artificial features and naturally matches the sensor model of a rotating 3D LiDAR. This preserves accurate neighbor relations along the edges, which would otherwise be compromised during the projection to the 2D range image. The Circular Padding (CP) also allows us to fit flexible input sizes without changing the patch geometry.

\subsubsection{Non-Square window for local self-attention}
Inside Swin Transformer, input feature maps are window-partitioned along width and height to compute the self-attention locally in each window. In addition to square-shaped RGB images, some recent work applies a square window for panoramic data as well~\cite{ling2023panoswin, yun2022panovit}. However, using square windows on range images is unfavorable. On one hand, small windows can focus more on attention within near areas but lead to increasing computational complexity and can struggle with the reconstruction of objects at a further distance in the scene.
On the other hand, a larger square window can blend scene-related contexts at different scales, which can potentially degrade the network performance. In a prior study done by Eskandar \etal{}~\cite{eskandar2022hals}, they demonstrated a significant enhancement in range image super-resolution by separately upsampling the upper and lower portions of the data. In contrast to their approach of utilizing CNN-based shallower and deeper network branches for distinct feature extraction, our proposal involves the direct partitioning of the input using a rectangular window. This window geometry facilitates robust attention interaction among distant points through W-MSA and subtle cross-attention between objects at multiple scales via SW-MSA and hierarchical processing. In addition, range images, as typical panorama data, inherently contain abundant information along the horizontal direction. Therefore, directing attention more toward image width rather than height can aid the network in capturing a greater amount of geometrical information. At the same time, the computational complexity remains the same as using a large square window.

\subsubsection{Further Adaptation and Refinement}
\PAR{Patch Unmerging (PU) and Pixel Shuffle (PS):} Similar to the patch merging layer that is responsible for downsampling at the encoder stage, a mechanism is necessary within the decoder to upsample the feature maps. Conventional CNN-based approaches~\cite{wang2022uformer, shan2020simulation} achieve upsampling through the transposed convolutional operation. As they tend to smooth out image sharpness~\cite{liang2021swinir}, we reversed the operation of patch merging, re-arranging the channels of patches into their respective positions in a grid to upsample the spatial resolution of input feature maps, which we denote as patch unmerging.
Furthermore, to reconstruct the final range image, we built the reconstruction head upon the upsampled feature maps. The implementation of the reconstruction module is based on the pixel shuffle layer~\cite{shi2016pixelshuffle}.
\PAR{Monte Carlo Dropout:} We additionally use Monte Carlo Dropout~\cite{shan2020simulation} to filter unreliable points by thresholding the uncertainty. MC-Dropout executes several feed-forward passes during inference with different active dropouts, which yields a distribution of outputs. We refine the results by removing points with high output variance, as they often come down to noisy, invalid points.

\section{Experiments}
\label{sec:experiments}
 \subsection{Experimental Settings}
 \PAR{Datasets:} We conduct experiments on three different datasets, that include the two large-scale real-world datasets DurLAR~\cite{li21durlar} and KITTI~\cite{geiger2012KITTI} (Sec. \ref{sec:real}), as well as on a dataset~\cite{kwon2022implicit} that was generated using the CARLA simulator~\cite{dosovitskiy2017carla} (Sec. \ref{sec:sim}). We select test sequences that are recorded in different locations than the train set to avoid spatial overlap. We vertically downsample the high-resolution range images with a factor of four by skipping the respective lines to generate low-resolution input images. For ablation studies in Sec.~\ref{sec:ablation}, we select the KITTI dataset.
 \PAR{Implementation Details:} In all experiments, we use distributed processing with $4\times$ GeForce RTX 2080 Ti. For training, we use a fixed batch size of eight per GPU for all datasets. We use the AdamW~\cite{adamw} optimizer with a weight decay factor of 0.01 and a base learning rate of 5e-4.
\PAR{Evaluation metrics:} We evaluate the Mean Absolute Error (MAE) for all the pixels in the generated 2-D range images. Additionally, we assess the performance by considering the 3D points reconstructed by our neural networks. The Chamfer Distance (CD) measures the Euclidean distance between two point clouds. We also follow the approach in \cite{kwon2022implicit} to evaluate the volumetric occupancy similarity. To do so, the point clouds are voxelized using a voxel size of $0.1m$. A voxel is classified as occupied for each point cloud if it contains at least one point. We then calculate the Intersection-over-Unit (IoU) based on the occupancy.

 \subsection{Ablation Studies} \label{sec:ablation}

\begin{table}[b!]
  \vspace{-2mm}
  \footnotesize
  \addtolength{\tabcolsep}{-4.2pt} 
  \begin{tabular}{c|cccc|ccc}
  Model & Patch & Window & CP/PU/PS & MC & MAE $\downarrow$ & IoU $\uparrow$ & CD $\downarrow$ \\
  \hline
  Baseline~\cite{cao2022swinunet} & $4\times4$ & Square &  $\times$ & $\times$ & 0.7138 &0.3250 &0.1940\\
  Model 1 & $2\times2$ & Square  & $\times$ & $\times$ & 0.6251 & 0.3337 &  0.1661 \\
  Model 2 & $1\times4$ & Square  &  $\times$ & $\times$ & 0.4814 & 0.3667 & 0.1391  \\
  Model 3 & $1\times4$ & Rectangular  & $\times$ & $\times$ &  0.4248 & 0.4040 & 0.1218 \\
  Model 4 & $1\times4$ & Rectangular &  $\checkmark$ & $\times$  & 0.4227& 0.4084 &0.1207  \\
  \coolname{} & $1\times4$ & Rectangular  & $\checkmark$ & $\checkmark$ & 0.4185 &0.4174&0.1207  \\
  \coolname{}-L & $1\times4$ & Rectangular & $\checkmark$ & $\checkmark$ & 0.3708 & 0.4329 &  0.0992\\
  \arrayrulecolor[rgb]{0.5,0.5,0.5} \hline
  ViT-Unet* &  $1\times4$ & - & $\checkmark$ & $\checkmark$ & 1.4484 & 0.2948 & 0.2665\\
  \end{tabular}
    \addtolength{\tabcolsep}{3.5pt} 
  \caption{Ablation study results for $4\times$ upsampling on KITTI. Input Resolution: $16\times1024$, Output Resolution: $64\times1024$. *We replaced Swin Transformer blocks with ViT blocks and followed the same training procedure as for \coolname{}.} %
  \label{Tab4:ablation}
  \vspace{-5pt}
\end{table}

\begin{figure}[t!]
  \centering
  \begin{subfigure}[b]{0.13\textwidth}
  \centering
    \includegraphics[width=\textwidth]{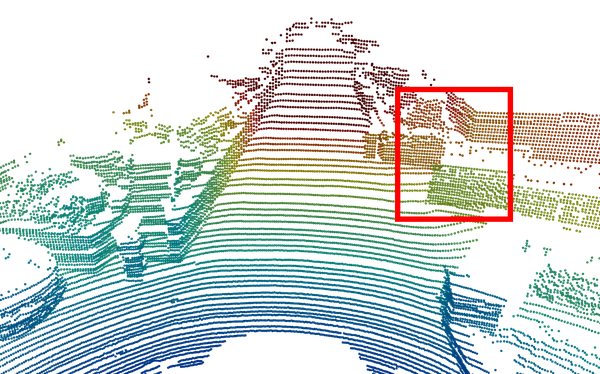}
    \caption{Scene}
  \end{subfigure}
  \begin{subfigure}[b]{0.08\textwidth}
    \includegraphics[width=\textwidth]{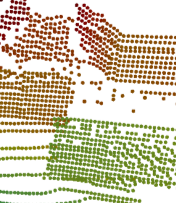}
    \caption{GT}
  \end{subfigure}
  \begin{subfigure}[b]{0.08\textwidth}
    \includegraphics[width=\textwidth]{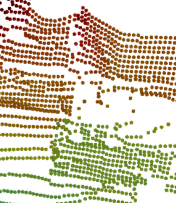}
    \caption{$1\times4$}
  \end{subfigure}
  \begin{subfigure}[b]{0.08\textwidth}
    \includegraphics[width=\textwidth]{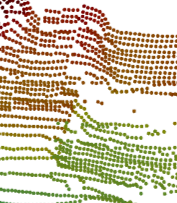}
    \caption{$2\times2$}
  \end{subfigure}
  \begin{subfigure}[b]{0.08\textwidth}
    \includegraphics[width=\textwidth]{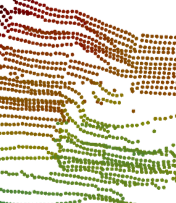}
    \caption{$4\times4$}
  \end{subfigure}
  \vspace{-2mm}
  \caption{Upsampling results with different patch sizes: We observe that square patches ($4\times4$, $2\times2$) blur out edges and create invalid connections between separate walls, while the proposed $1\times4$ patch generates sharp edges and correct discontinuities.}
  \vspace{-6mm}
  \label{fig:exp_patch_size}
\end{figure}

\PAR{Patch size:} In Tab.~\ref{Tab4:ablation} Baseline and Model 1-2, we compare the upsampling results with different patch sizes. We observe that the smaller patch ($2 \times 2$) outperforms the larger one ($4 \times 4$) while the proposed row patch ($1\times4$) can further refine the network performance significantly. In particular, the row patch improves $32.6\%$ in MAE, $12.8\%$ in IoU, and $28.3\%$ in CD, compared to a $4 \times 4$ patch. As visualized in Fig.~\ref{fig:exp_patch_size}, the row patch better preserves sharpness around edges and corners compared to the larger patch sizes.

\begin{figure}[!]
  \centering
  \begin{subfigure}[b]{0.15\textwidth}
  \centering
    \includegraphics[width=\textwidth]{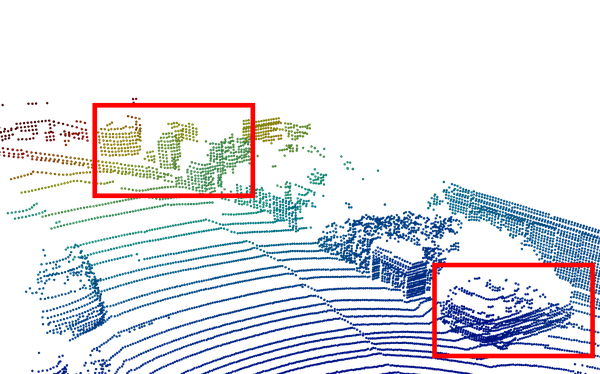}
    \caption{Scene}
  \end{subfigure}
  \begin{subfigure}[b]{0.1\textwidth}
    \includegraphics[width=\textwidth]{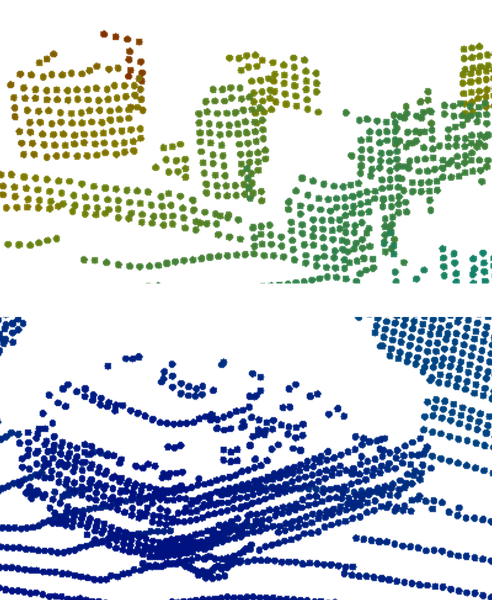}
    \caption{GT}
  \end{subfigure}
  \begin{subfigure}[b]{0.1\textwidth}
    \includegraphics[width=\textwidth]{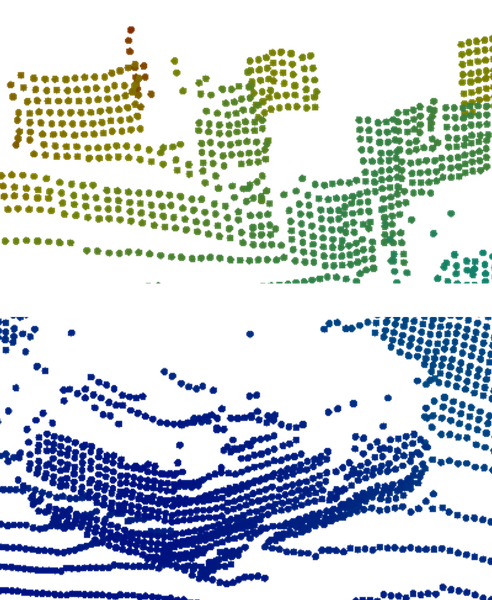}
    \caption{Rectangular}
  \end{subfigure}
  \begin{subfigure}[b]{0.1\textwidth}
    \includegraphics[width=\textwidth]{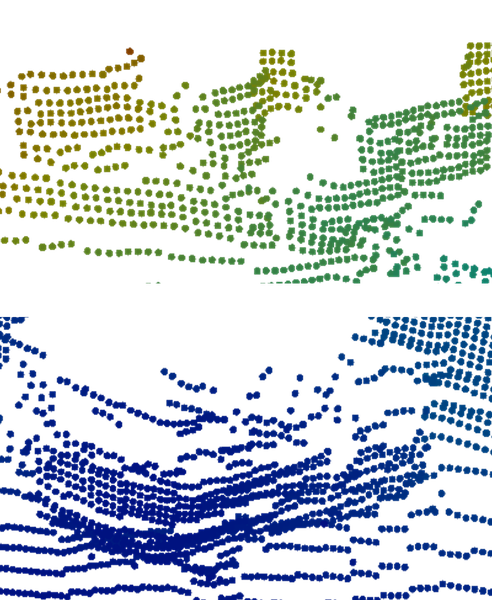}
    \caption{Square}
  \end{subfigure}
  \vspace{-3mm}
  \caption{Upsampling results with different window geometries: Rectangular windows improve reconstruction of objects at different ranges, indicated by more distinct discontinuities (top) and a better reconstruction of the car (bottom).}
  \label{fig:exp_window_size}
  \vspace{-6mm}
\end{figure}

\PAR{Window geometry:} To verify the effectiveness of modifying window geometry for range images, we conduct experiments with square and rectangular windows (Tab.~\ref{Tab4:ablation} Model 2-3. Based on the quantitative results, we can observe the following: reconstruction quality produced by local attention within large square windows is inferior to the rectangular window concerning both 2D and 3D evaluation metrics, while they result in the same computational complexity. The underlying reason for this performance difference lies in the perceptive field of the windows. A wider window is more likely to include discontinuities between objects at different scales. Accordingly, the network performs better at separating objects from each other, which can visually be validated by clearer surface boundaries in Fig.~\ref{fig:exp_window_size}.

\PAR{Further Adaptation and Refinement:} Circular Padding (CP) helps maintain continuity and consistency at the edges of the panoramic images, but it has minor contributions to other areas. Patch Unmerging (PU), and Pixel Shuffle (PS) aim at upscaling the spatial resolution with rearrangement, avoiding information loss and parameter increase led by using additional de-convolutional layers. Although they are designed to improve efficiency rather than efficacy of the model, tested with those components (Model 4), a slight improvement in upsampling is still observable. Monte Carlo Dropout~\cite{gal2016dropout}, as a post-processing step, helps to further reduce ghost points in between objects and we present more details in the Appendix. For \coolname{}-L, we increase the model capacity, using four Swin-Transformer layers instead of three in both encoder and decoder, which demonstrates an additional performance boost on all metrics.

\PAR{Transformer Block:} From the previous discussion, we infer that Swin Transformer~\cite{liu2021swintransformer} is more advantageous than ViT (Vision Transfromer)~\cite{dosovitskiy2020vit} in LiDAR upsampling. On one hand, prior works~\cite{gani2022howvit, lee2021vision, liu2021efficient, dai2023swinmae} have pointed out the inferiority of ViT on smaller datasets due to lack of locality learning and non-overlapping attention. Relevant LiDAR datasets~\cite{geiger2012KITTI, li21durlar} are generally small compared to RGB dataset~\cite{russakovsky2015imagenet}. On the other hand, by narrowing the patch size, the exponential increase of model capacity hinders sufficient training with ViT on range images, while Swin Transformer expands only linearly. To numerically show the distinction, we trained the network on the same dataset by replacing ViT blocks and applied the same patch size and additional components to the network for a fair comparison. In the last row of Tab.~\ref{Tab4:ablation}, it shows that \coolname{} significantly outperforms the one with ViT as the backbone.
\subsection{Benchmark Results}
\begin{table}[t]
\centering
\renewcommand \arraystretch{1.1}
\begin{tabular*}{\linewidth}{@{\extracolsep{\fill}} l c c c }
\hline
Model & MAE $\downarrow$ & IoU $\uparrow$ & CD $\downarrow$\\
\hline
\multicolumn{4}{c}{CARLA 4x Output Resolution: $128 \times 2048$} \\
\hline
Bilinear & 1.8128  & 0.1382 & 0.7262\\
SRNO~\cite{wei2023srno} & 2.4640 &  0.1343 & 2.0230\\
HAT~\cite{chen2023hat}  & 1.6032 & 0.2698 & 0.6337\\
SWIN-IR~\cite{liang2021swinir} & 1.9560 & 0.2718 &0.4840\\
LIIF~\cite{chen2021liif} & 0.8064 &  0.3502 & 0.174\\
\arrayrulecolor[rgb]{0.8,0.8,0.8} \hline
LIDAR-SR~\cite{shan2020simulation} & 0.8216 & 0.2581  & 0.2044\\
ILN~\cite{kwon2022implicit} & 0.8592 & 0.5006 & 0.1855\\
\coolname{} (Ours) & \underline{0.7699} & \underline{0.5152} & \underline{0.1028}\\
\rowcolor{green!15}\coolname{}-L (Ours) & \textbf{0.7539}  & \textbf{0.5301} & \textbf{0.1001} \\
\arrayrulecolor{black}\hline
\multicolumn{4}{c}{KITTI 4x Output Resolution: $64 \times 1024$} \\
\hline
Bilinear & 2.0892 & 0.1063 & 0.6000\\
SRNO~\cite{wei2023srno} & 0.8350 & 0.2035 & 0.4417\\
HAT~\cite{chen2023hat}  & 0.6856 & 0.2035 & 0.2516\\
SWIN-IR~\cite{liang2021swinir} & 1.2972 & 0.2774 & 0.7347\\
LIIF~\cite{chen2021liif} & 0.6143 & 0.3226 & 0.1916\\
\arrayrulecolor[rgb]{0.8,0.8,0.8} \hline
LIDAR-SR~\cite{shan2020simulation} & 0.5674 & 0.1005 & 0.2165\\
ILN~\cite{kwon2022implicit} & 1.0528  & 0.3342 & 0.2787\\
\coolname{} (Ours) & \underline{0.4185} & \underline{0.4174} & \underline{0.1207}\\
\rowcolor{green!15}\coolname{}-L (Ours) & \textbf{0.3708} & \textbf{0.4329} & \textbf{0.0992}\\
\arrayrulecolor{black}\hline 
\multicolumn{4}{c}{DurLAR 4x Output Resolution: $128 \times 2048$} \\
\hline
Bilinear & 2.4384  & 0.1266 & 0.6346\\
SRNO~\cite{wei2023srno} & 1.5396 & 0.1507 & 0.5108\\
HAT~\cite{chen2023hat}  & 1.7820 & 0.2353 & 0.1973\\
SWIN-IR~\cite{liang2021swinir} &  1.9416  & 0.2157 & 0.2279\\
LIIF~\cite{chen2021liif} & 1.5672 & 0.2469 & 0.1548\\
\arrayrulecolor[rgb]{0.8,0.8,0.8} \hline
LIDAR-SR~\cite{shan2020simulation} & \textbf{1.5312} & 0.1370 & 0.1128\\
ILN~\cite{kwon2022implicit} & 1.5720 & 0.3430 & 0.0893\\
\coolname{} (Ours) & \underline{1.5432} & \underline{0.3562} & \underline{0.06484}\\
\rowcolor{green!15}\coolname{}-L (Ours) & 1.5592 & \textbf{0.3654} &\textbf{0.06346} \\
\arrayrulecolor{black}\hline
\end{tabular*}
\vspace{-1em}
\caption{Quantitative comparison against state-of-the-art LiDAR and image super-resolution methods on different datasets. All methods are trained and evaluated on the same splits.}
\label{tab:main}
\vspace*{-\baselineskip}
\end{table}
Besides bilinear interpolation, which computes the interpolation weights from the four nearest neighbors, we evaluate against the state-of-the-art LiDAR upsampling approaches Implicit LiDAR Network (ILN)~\cite{kwon2022implicit} and LiDAR-SR~\cite{shan2020simulation}. We also compare against several image super-resolution works: SRNO ~\cite{wei2023srno}, and LIIF~\cite{chen2021liif} learn implicit features for interpolation instead of 2D coordinates of new pixels while HAT~\cite{chen2023hat} and Swin-IR~\cite{liang2021swinir} are pixel-based approaches that directly reconstruct high-resolution images from low-resolution feature embeddings. Similar to our work, they are based on Swin Transformer~\cite{liu2021swintransformer}. We train all methods on the specific datasets using their default parameters.
\begin{figure}[b]
    \centering
    \begin{tabular*}{\linewidth}{@{\extracolsep{\fill}}m{-1cm}m{2cm}m{2cm}m{2cm}}
    \rotatebox{90}{\fontsize{7}{10}\selectfont Input}&\includegraphics[width=0.14\textwidth]{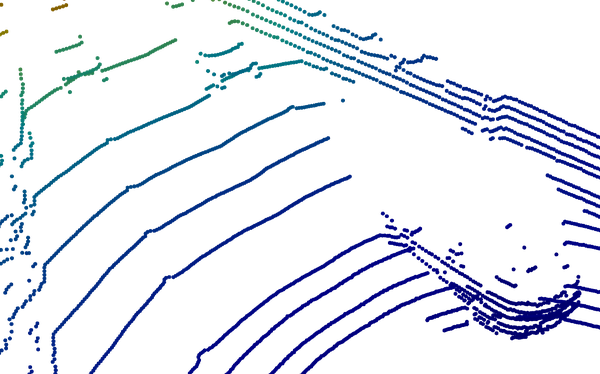}&\includegraphics[width=0.14\textwidth]{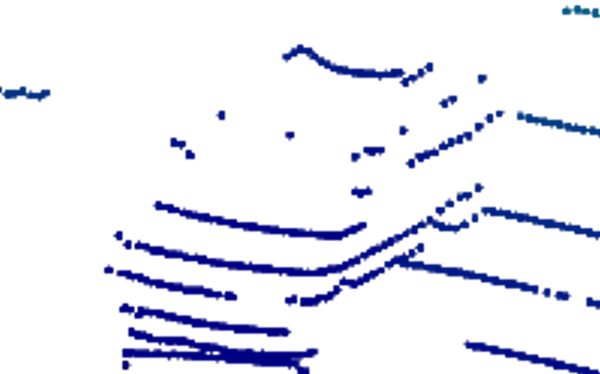}&\includegraphics[width=0.14\textwidth]{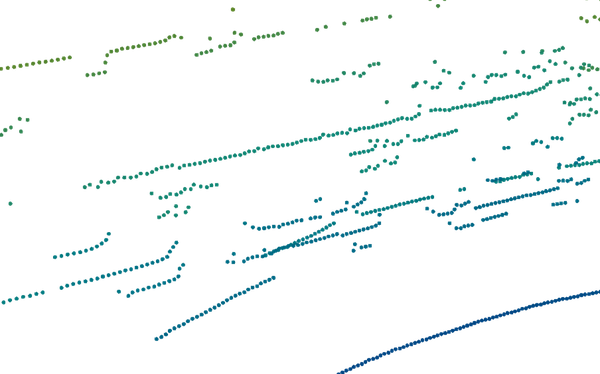}\\
    \rotatebox{90}{\fontsize{7}{10}\selectfont Swin-IR~\cite{liang2021swinir}}&\includegraphics[width=0.14\textwidth]{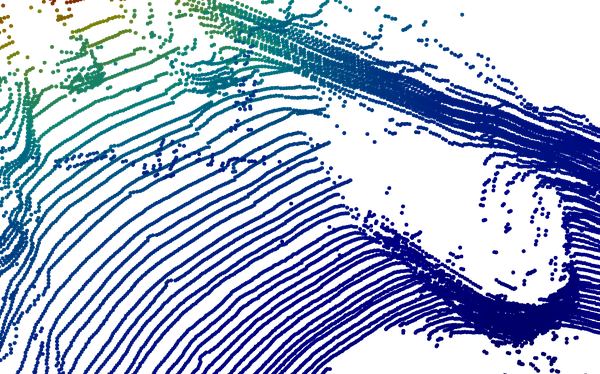}&\includegraphics[width=0.14\textwidth]{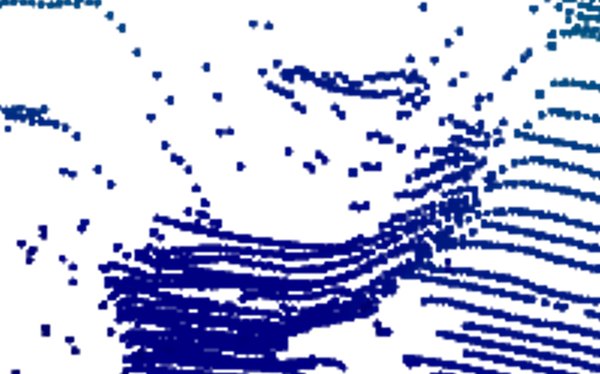}&\includegraphics[width=0.14\textwidth]{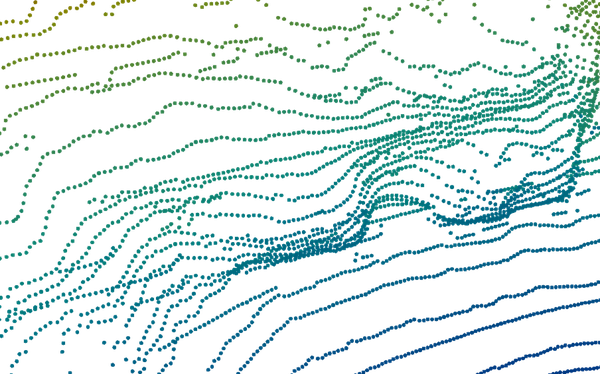 }\\
    \rotatebox{90}{\fontsize{7}{10}\selectfont LIIF~\cite{chen2021liif}}&\includegraphics[width=0.14\textwidth]{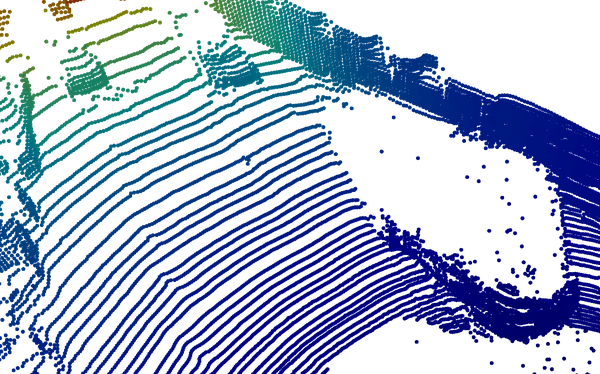}&\includegraphics[width=0.14\textwidth]{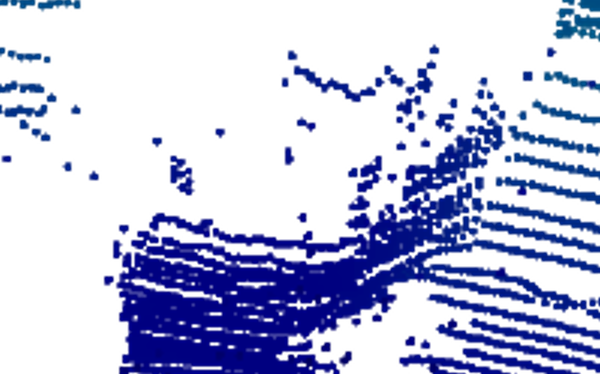}&\includegraphics[width=0.14\textwidth]{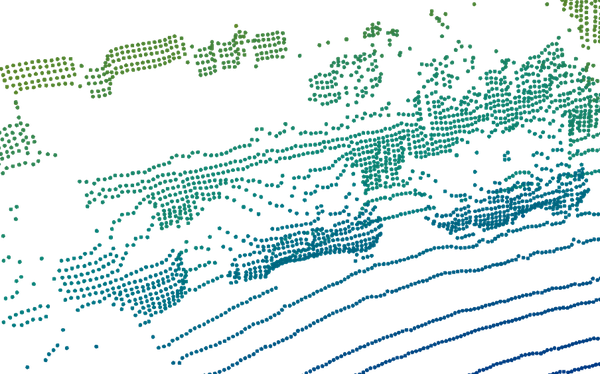}\\
    \rotatebox{90}{\fontsize{7}{10}\selectfont LiDAR-SR~\cite{shan2020simulation}}&\includegraphics[width=0.14\textwidth]{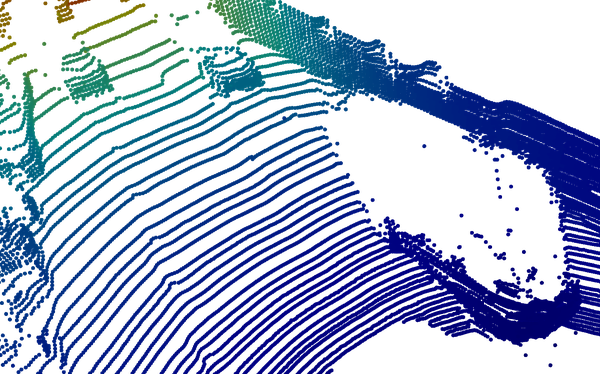}&\includegraphics[width=0.14\textwidth]{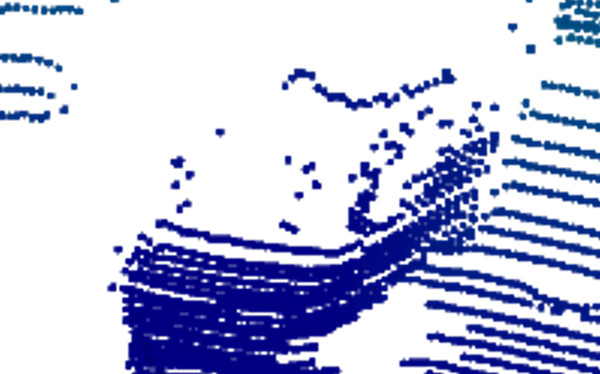}&\includegraphics[width=0.14\textwidth]{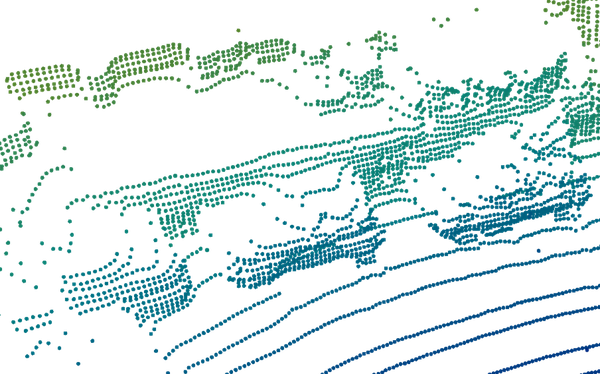}\\
    \rotatebox{90}{\fontsize{7}{10}\selectfont ILN~\cite{kwon2022implicit}}&\includegraphics[width=0.14\textwidth]{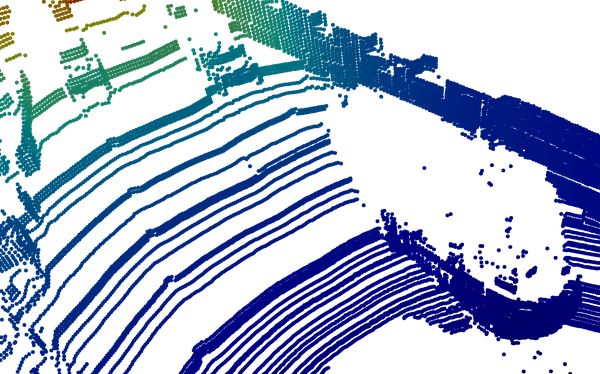}&\includegraphics[width=0.14\textwidth]{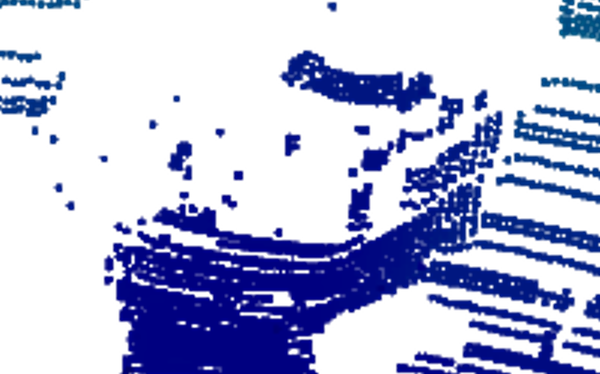}&\includegraphics[width=0.14\textwidth]{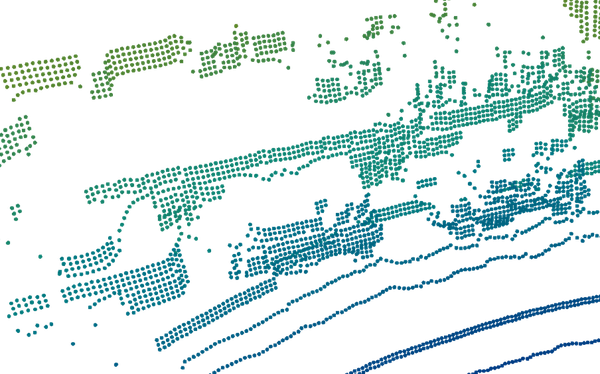}\\
    \rotatebox{90}{\fontsize{7}{10}\selectfont \coolname{} (Ours)} &\includegraphics[width=0.14\textwidth]{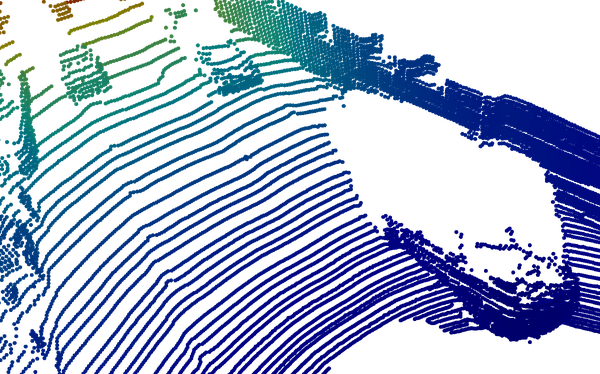}&\includegraphics[width=0.14\textwidth]{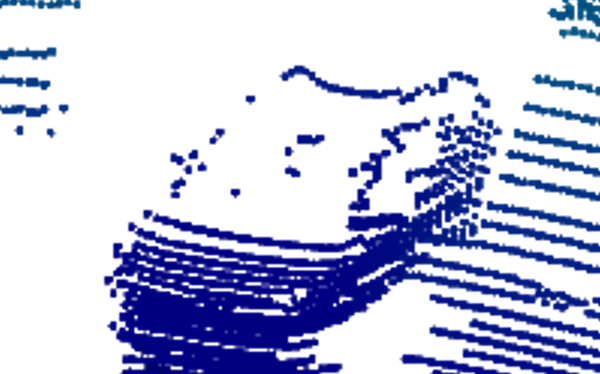}&\includegraphics[width=0.14\textwidth]{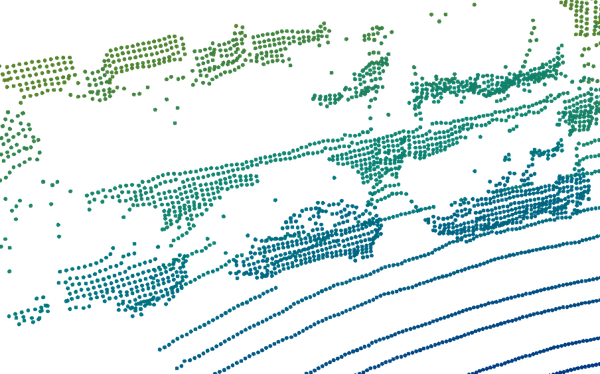}\\
    \rotatebox{90}{\fontsize{7}{10}\selectfont GT}&\includegraphics[width=0.14\textwidth]{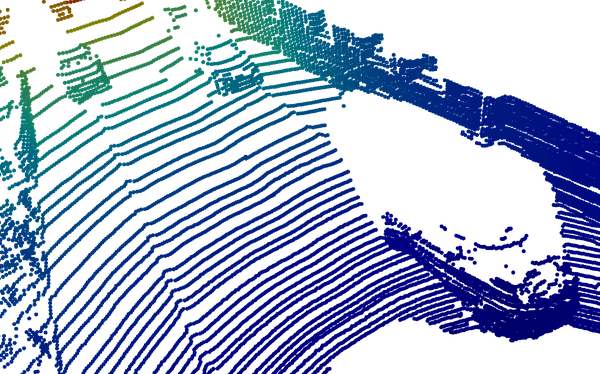}&\includegraphics[width=0.14\textwidth]{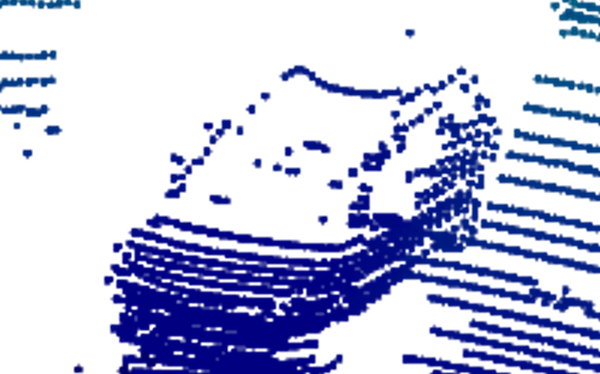}&\includegraphics[width=0.14\textwidth]{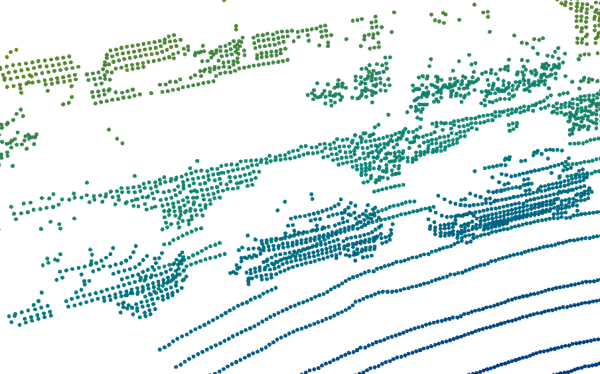}\\
    \quad & \hspace{10mm} A & \hspace{10mm} B & \hspace{10mm} C\\
    
    \end{tabular*}
    
    \vspace{-1mm}
    \caption{Qualitative results on KITTI: We provide visualizations for the top-4 results from Tab.~\ref{tab:main} and Swin-IR~\cite{liang2021swinir}. Our approach outperforms other state-of-the-art methods in upsampling realistic point clouds. \textbf{A}: \coolname{} is the only approach that does not generate invalid ghost points in between the car (right bottom) and the wall behind. It can also be observed that all approaches except for ILN match the characteristic line pattern of the LiDAR on the ground (middle). \textbf{B}: Swin-IR and LIIF generate a blurry reconstruction. \coolname{} achieves the clearest point cloud and even reconstructs small details such as the car's side mirrors. \textbf{C}: Swin-IR, LIIF, and LiDAR-SR do not clearly separate the car from the wall behind. ILN generates a clear point cloud but does not properly reconstruct the shape of the cars. \coolname{} achieves the clearest point cloud, with distinct discontinuities between objects, and closely resembles the shape of the three cars.}
    \vspace{-6mm}
    \label{fig:vis}
\end{figure}

\subsubsection{Simulation Results} \label{sec:sim}
Similar to prior works, we first evaluate the performance on noise-free, simulated data. We train and test all approaches on the dataset presented in ~\cite{kwon2022implicit} that was recorded using the CARLA simulator and select the ground-truth range images of size $128\times2048$ that capture a vertical FoV of 30$^\circ$. We use a (20699/2618) train/test split. We observe that the performance between image super-resolution approaches varies greatly, and LIIF performs the best among them. While LIIF~\cite{chen2021liif} achieves a comparable MAE to the LiDAR upsampling methods, the 3D metrics are significantly worse, indicating that it is unsuitable for the underlying geometric input data. Nevertheless, LIIF~\cite{chen2021liif} still outperforms the LiDAR upsampling method in LIDAR-SR. The interpolation-based approach in ILN shows decent results for the noise-free data, as indicated by their high IoU. Nevertheless, our approaches achieve the best performance in all evaluated metrics, with the strongest improvement in Chamfer Distance, which validates that our design is well suited for the geometric reconstruction in range images.

\subsubsection{Real-World Results} \label{sec:real}
To evaluate our approach on real-world data, we downsample the DurLAR and KITTI datasets to a comparable number of frames to the CARLA dataset by temporally skipping frames in the dataset sequences. In particular, we have the following (train/test) split: DurLAR (24372/2508), and KITTI (20000/2500).

\PAR{KITTI:} The KITTI dataset~\cite{geiger2012KITTI} was collected using a Velodyne HDL-64E LiDAR with a vertical FoV of 26.8$^\circ$ and a resolution of $64\times1024$.
For the image super-resolution approaches, we observe comparable results and trends on KITTI as on the simulated data in Sec.\ref{sec:sim}. Even though LIDAR-SR~\cite{shan2020simulation} is designed for LIDAR upsampling and outperforms image super-resolution approaches in terms of MAE, it achieves the lowest IoU among all approaches, which can be explained by a large amount of invalid floating points between real objects. LIIF~\cite{chen2021liif} achieves better CD and comparable IoU compared to ILN. We observe that the IoU of ILN~\cite{kwon2022implicit} is drastically lower than on CARLA. The interpolation scheme in ILN is thus not suitable to properly handle the lower resolution data with noise from the real-world sensor. Our approaches significantly outperform all competing methods, especially in terms of IoU as it can better handle discontinuities between objects. As indicated in the visualizations in Fig.\ref{fig:vis}, our approach predicts fewer invalid float points between objects, as well as fewer blurry edges, which results in much clearer point clouds.

\PAR{DurLAR:} The DurLAR dataset~\cite{li21durlar} was recorded using an Ouster OS1-128 LiDAR, that captures point clouds with a resolution of $128\times2048$ with a 45$^\circ$ vertical FoV. 
Compared to CARLA and KITTI, we see a large performance drop for all approaches in MAE and IoU, which results from the larger maximum range and vertical FoV of the sensor as well as stronger sensor noise. We observe that all approaches designed for LiDAR upsampling clearly improve in Chamfer Distance compared to image super-resolution approaches. ILN and \coolname{} outperform all methods by a large margin in IoU. \coolname{}-L achieves the best 3D metrics, which validates that it can also handle the challenging DurLAR data better than prior works.

\PAR{Range Analysis:} As the density of LiDAR point clouds decreases with increasing distance from the sensor, we additionally compare the performance at different ranges for a more fine-grained evaluation. We observe that our approach achieves better or similar performance compared to all approaches in all metrics at all ranges. Up to 30m, \coolname{} clearly outperforms the baseline approaches. \coolname{} handles discontinuities between objects better than other approaches, which can be verified visually through sharper edges and fewer "floating" points between objects. This is especially reflected in the high IoU. 
At ranges above 30m, our approach performs comparable to other methods. Above this range, LiDAR data is extremely sparse, meaning that points neighboring in the low-resolution range image are actually distant in the Euclidean space. It is, therefore, not feasible to infer meaningful information about the scene from such neighboring points at a high range, which is also represented in the poor performance of all approaches.

\begin{figure}[t!]
  \centering
  \begin{subfigure}[b]{0.2335\textwidth}
    \includegraphics[width=\textwidth]{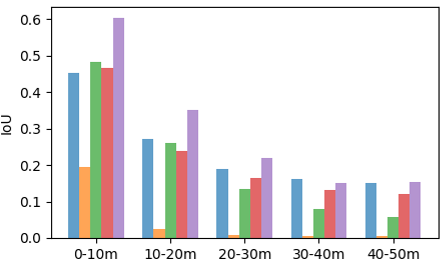}
    \caption{IoU $\uparrow$}
  \end{subfigure}
  \begin{subfigure}[b]{0.235\textwidth}
    \includegraphics[width=\textwidth]{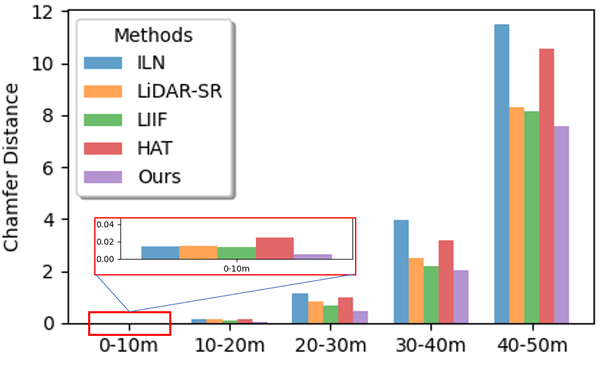}
    \caption{Chamfer distance $\downarrow$}
  \end{subfigure}
  \vspace{-5mm}
  \caption{3D error metrics visualized at different ranges for KITTI.}
  \vspace{-6mm}
  \label{fig:exp_range_ablation}
\end{figure}

\subsection{Failure Cases}
\begin{figure}[h] %
    \hspace{0.5cm}
    \begin{subfigure}[b]{0.22\textwidth}
    \includegraphics[width=\textwidth]{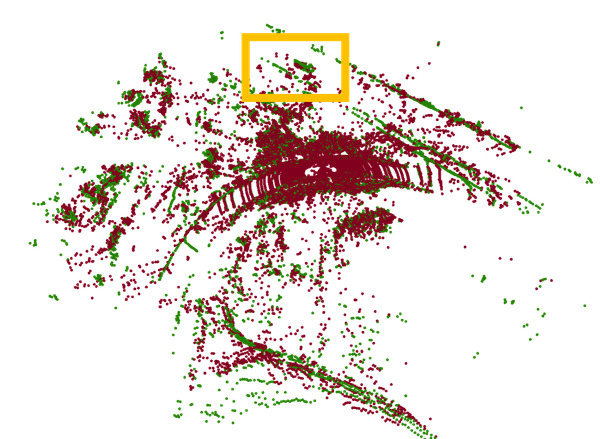}
    \caption{Full scene}
    \label{fig:fullscene_failurecase}
  \end{subfigure}
  \hspace{0.5cm}
  \begin{subfigure}[b]{0.15\textwidth}
    \includegraphics[width=\textwidth]{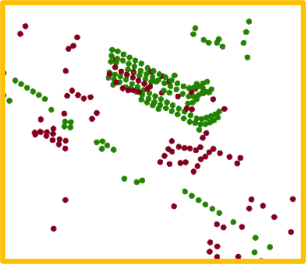}
    \caption{Details}
  \end{subfigure}
  \caption{Failure Case: noisy point cloud generated by \coolname{} and missing reconstruction of the object. (\textcolor{red}{Ours} and \textcolor{green}{GT})}
  \label{fig:failue_case}
\end{figure}

Although \coolname{} outperforms other state-of-the-art methods both qualitatively and quantitatively, the upsampling quality can still be limited in certain cases. For instance, as shown in Fig~\ref{fig:failue_case}, our approach shows inferior upsampling results for the specific scene. The irregularity of the scene leads to high uncertainty, which makes the reconstruction noisy. Observing the details in Fig.~\ref{fig:fullscene_failurecase}, the network fails to reconstruct the car at the top of the scene.

\vspace{-3mm}

\section{Conclusion}
\label{sec:conclusion}
This work presents \coolname{}, a novel method for LiDAR Upsampling that achieves incredible performance in upsampling the range image. Our approach transforms 3D point clouds into 2D range images and performs the upsampling in 2D space. We build upon a Swin-Transformer-based network and specifically modify the patch partition and attention windows to better accommodate the characteristics of range images. Throughout various experiments, testing on three different benchmarks, it shows that \coolname{} outperforms state-of-the-art methods quantitatively in all evaluation metrics, and quantitatively, \coolname{} demonstrates more promising results in upsampling the realistic LiDAR data.

{\small
\textbf{Acknowledgement:} This work was supported by Swiss National Science Foundation’s NCCR DFab P3.
}
\label{sec:ackowledgement}

{
    \small
    \bibliographystyle{ieeenat_fullname}
    \bibliography{main}

\begin{thebibliography}{66}
\providecommand{\natexlab}[1]{#1}
\providecommand{\url}[1]{\texttt{#1}}
\expandafter\ifx\csname urlstyle\endcsname\relax
  \providecommand{\doi}[1]{doi: #1}\else
  \providecommand{\doi}{doi: \begingroup \urlstyle{rm}\Url}\fi

\bibitem[Akhtar et~al.(2022)Akhtar, Li, Van~der Auwera, Li, and Chen]{akhtar2022pu}
Anique Akhtar, Zhu Li, Geert Van~der Auwera, Li Li, and Jianle Chen.
\newblock Pu-dense: Sparse tensor-based point cloud geometry upsampling.
\newblock \emph{IEEE Transactions on Image Processing}, 31:\penalty0 4133--4148, 2022.

\bibitem[Amir et~al.(2022)Amir, Gandelsman, Bagon, and Dekel]{amir2021deepvit}
Shir Amir, Yossi Gandelsman, Shai Bagon, and Tali Dekel.
\newblock Deep vit features as dense visual descriptors.
\newblock \emph{ECCVW What is Motion For?}, 2022.

\bibitem[Be{\v{s}}i{\'c} et~al.(2022)Be{\v{s}}i{\'c}, Gosala, Cattaneo, and Valada]{bevsic2022unsupervised}
Borna Be{\v{s}}i{\'c}, Nikhil Gosala, Daniele Cattaneo, and Abhinav Valada.
\newblock Unsupervised domain adaptation for lidar panoptic segmentation.
\newblock \emph{IEEE Robotics and Automation Letters}, 7\penalty0 (2):\penalty0 3404--3411, 2022.

\bibitem[Brown et~al.(2020)Brown, Mann, Ryder, Subbiah, Kaplan, Dhariwal, Neelakantan, Shyam, Sastry, Askell, et~al.]{brown2020language}
Tom Brown, Benjamin Mann, Nick Ryder, Melanie Subbiah, Jared~D Kaplan, Prafulla Dhariwal, Arvind Neelakantan, Pranav Shyam, Girish Sastry, Amanda Askell, et~al.
\newblock Language models are few-shot learners.
\newblock \emph{Advances in neural information processing systems}, 33:\penalty0 1877--1901, 2020.

\bibitem[Cao et~al.(2022)Cao, Wang, Chen, Jiang, Zhang, Tian, and Wang]{cao2022swinunet}
Hu Cao, Yueyue Wang, Joy Chen, Dongsheng Jiang, Xiaopeng Zhang, Qi Tian, and Manning Wang.
\newblock Swin-unet: Unet-like pure transformer for medical image segmentation.
\newblock In \emph{European conference on computer vision}, pages 205--218. Springer, 2022.

\bibitem[Carion et~al.(2020)Carion, Massa, Synnaeve, Usunier, Kirillov, and Zagoruyko]{carion2020detr}
Nicolas Carion, Francisco Massa, Gabriel Synnaeve, Nicolas Usunier, Alexander Kirillov, and Sergey Zagoruyko.
\newblock End-to-end object detection with transformers.
\newblock In \emph{European conference on computer vision}, pages 213--229. Springer, 2020.

\bibitem[Caron et~al.(2021)Caron, Touvron, Misra, J{\'e}gou, Mairal, Bojanowski, and Joulin]{caron2021dino}
Mathilde Caron, Hugo Touvron, Ishan Misra, Herv{\'e} J{\'e}gou, Julien Mairal, Piotr Bojanowski, and Armand Joulin.
\newblock Emerging properties in self-supervised vision transformers.
\newblock In \emph{Proceedings of the IEEE/CVF international conference on computer vision}, pages 9650--9660, 2021.

\bibitem[Chen et~al.(2022)Chen, Hsiao, and Huang]{chen2022density}
Tso~Yuan Chen, Ching~Chun Hsiao, and Ching-Chun Huang.
\newblock Density-imbalance-eased lidar point cloud upsampling via feature consistency learning.
\newblock \emph{IEEE Transactions on Intelligent Vehicles}, 2022.

\bibitem[Chen et~al.(2021{\natexlab{a}})Chen, Vizzo, L{\"a}be, Behley, and Stachniss]{chen2021range}
Xieyuanli Chen, Ignacio Vizzo, Thomas L{\"a}be, Jens Behley, and Cyrill Stachniss.
\newblock Range image-based lidar localization for autonomous vehicles.
\newblock In \emph{International Conference on Robotics and Automation (ICRA)}, pages 5802--5808, 2021{\natexlab{a}}.

\bibitem[Chen et~al.(2023{\natexlab{a}})Chen, Wang, Zhou, Qiao, and Dong]{chen2023hat}
Xiangyu Chen, Xintao Wang, Jiantao Zhou, Yu Qiao, and Chao Dong.
\newblock Activating more pixels in image super-resolution transformer.
\newblock In \emph{Proceedings of the IEEE/CVF Conference on Computer Vision and Pattern Recognition}, pages 22367--22377, 2023{\natexlab{a}}.

\bibitem[Chen et~al.(2021{\natexlab{b}})Chen, Liu, and Wang]{chen2021liif}
Yinbo Chen, Sifei Liu, and Xiaolong Wang.
\newblock Learning continuous image representation with local implicit image function.
\newblock In \emph{Proceedings of the IEEE/CVF Conference on Computer Vision and Pattern Recognition}, pages 8628--8638, 2021{\natexlab{b}}.

\bibitem[Chen et~al.(2023{\natexlab{b}})Chen, Zhang, Gu, Kong, Yang, and Yu]{chen2023dual}
Zheng Chen, Yulun Zhang, Jinjin Gu, Linghe Kong, Xiaokang Yang, and Fisher Yu.
\newblock Dual aggregation transformer for image super-resolution.
\newblock In \emph{Proceedings of the IEEE/CVF International Conference on Computer Vision}, pages 12312--12321, 2023{\natexlab{b}}.

\bibitem[Cheng et~al.(2018)Cheng, Wang, and Yang]{cheng2018depth}
Xinjing Cheng, Peng Wang, and Ruigang Yang.
\newblock Depth estimation via affinity learned with convolutional spatial propagation network.
\newblock In \emph{Proceedings of the European conference on computer vision (ECCV)}, pages 103--119, 2018.

\bibitem[Dai et~al.(2023)Dai, Liu, Chen, Liu, Shi, Liu, Zhou, et~al.]{dai2023swinmae}
Yin Dai, Fayu Liu, Weibing Chen, Yue Liu, Lifu Shi, Sheng Liu, Yuhang Zhou, et~al.
\newblock Swin mae: Masked autoencoders for small datasets.
\newblock \emph{Computers in Biology and Medicine}, 161:\penalty0 107037, 2023.

\bibitem[Dong et~al.(2015)Dong, Loy, He, and Tang]{dong2015imagesr}
Chao Dong, Chen~Change Loy, Kaiming He, and Xiaoou Tang.
\newblock Image super-resolution using deep convolutional networks.
\newblock \emph{IEEE transactions on pattern analysis and machine intelligence}, 38\penalty0 (2):\penalty0 295--307, 2015.

\bibitem[Dong et~al.(2016)Dong, Loy, and Tang]{dong2016accelerating}
Chao Dong, Chen~Change Loy, and Xiaoou Tang.
\newblock Accelerating the super-resolution convolutional neural network.
\newblock In \emph{Computer Vision--ECCV 2016: 14th European Conference, Amsterdam, The Netherlands, October 11-14, 2016, Proceedings, Part II 14}, pages 391--407. Springer, 2016.

\bibitem[Dosovitskiy et~al.(2017)Dosovitskiy, Ros, Codevilla, Lopez, and Koltun]{dosovitskiy2017carla}
Alexey Dosovitskiy, German Ros, Felipe Codevilla, Antonio Lopez, and Vladlen Koltun.
\newblock Carla: An open urban driving simulator.
\newblock In \emph{Conference on robot learning}, pages 1--16. PMLR, 2017.

\bibitem[Dosovitskiy et~al.(2021)Dosovitskiy, Beyer, Kolesnikov, Weissenborn, Zhai, Unterthiner, Dehghani, Minderer, Heigold, Gelly, et~al.]{dosovitskiy2020vit}
Alexey Dosovitskiy, Lucas Beyer, Alexander Kolesnikov, Dirk Weissenborn, Xiaohua Zhai, Thomas Unterthiner, Mostafa Dehghani, Matthias Minderer, Georg Heigold, Sylvain Gelly, et~al.
\newblock An image is worth 16x16 words: Transformers for image recognition at scale.
\newblock In \emph{International Conference on Learning Representations}, 2021.

\bibitem[Eskandar et~al.(2022)Eskandar, Sudarsan, Guirguis, Palaniswamy, Somashekar, and Yang]{eskandar2022hals}
George Eskandar, Sanjeev Sudarsan, Karim Guirguis, Janaranjani Palaniswamy, Bharath Somashekar, and Bin Yang.
\newblock Hals: A height-aware lidar super-resolution framework for autonomous driving.
\newblock \emph{arXiv preprint arXiv:2202.03901}, 2022.

\bibitem[Gal and Ghahramani(2016)]{gal2016dropout}
Yarin Gal and Zoubin Ghahramani.
\newblock Dropout as a bayesian approximation: Representing model uncertainty in deep learning.
\newblock In \emph{international conference on machine learning}, pages 1050--1059. PMLR, 2016.

\bibitem[Gani et~al.(2022)Gani, Naseer, and Yaqub]{gani2022howvit}
Hanan Gani, Muzammal Naseer, and Mohammad Yaqub.
\newblock How to train vision transformer on small-scale datasets?
\newblock In \emph{33rd British Machine Vision Conference}, 2022.

\bibitem[Geiger et~al.(2012)Geiger, Lenz, and Urtasun]{geiger2012KITTI}
Andreas Geiger, Philip Lenz, and Raquel Urtasun.
\newblock Are we ready for autonomous driving? the kitti vision benchmark suite.
\newblock In \emph{2012 IEEE conference on computer vision and pattern recognition}, pages 3354--3361. IEEE, 2012.

\bibitem[Hu et~al.(2021)Hu, Wang, Li, Ning, Fan, and Gong]{hu2021penet}
Mu Hu, Shuling Wang, Bin Li, Shiyu Ning, Li Fan, and Xiaojin Gong.
\newblock Penet: Towards precise and efficient image guided depth completion.
\newblock In \emph{2021 IEEE International Conference on Robotics and Automation (ICRA)}, pages 13656--13662. IEEE, 2021.

\bibitem[Kenton and Toutanova(2019)]{devlin2018bert}
Jacob Devlin Ming-Wei~Chang Kenton and Lee~Kristina Toutanova.
\newblock Bert: Pre-training of deep bidirectional transformers for language understanding.
\newblock In \emph{Proceedings of naacL-HLT}, page~2, 2019.

\bibitem[Kim et~al.(2016{\natexlab{a}})Kim, Lee, and Lee]{kim2016accurate}
Jiwon Kim, Jung~Kwon Lee, and Kyoung~Mu Lee.
\newblock Accurate image super-resolution using very deep convolutional networks.
\newblock In \emph{Proceedings of the IEEE conference on computer vision and pattern recognition}, pages 1646--1654, 2016{\natexlab{a}}.

\bibitem[Kim et~al.(2016{\natexlab{b}})Kim, Lee, and Lee]{kim2016deeply}
Jiwon Kim, Jung~Kwon Lee, and Kyoung~Mu Lee.
\newblock Deeply-recursive convolutional network for image super-resolution.
\newblock In \emph{Proceedings of the IEEE conference on computer vision and pattern recognition}, pages 1637--1645, 2016{\natexlab{b}}.

\bibitem[Kwon et~al.(2022{\natexlab{a}})Kwon, Kim, Mahoney, Hassoun, Keutzer, and Gholami]{kwon2022fast}
Woosuk Kwon, Sehoon Kim, Michael~W Mahoney, Joseph Hassoun, Kurt Keutzer, and Amir Gholami.
\newblock A fast post-training pruning framework for transformers.
\newblock \emph{Advances in Neural Information Processing Systems}, 35:\penalty0 24101--24116, 2022{\natexlab{a}}.

\bibitem[Kwon et~al.(2022{\natexlab{b}})Kwon, Sung, and Yoon]{kwon2022implicit}
Youngsun Kwon, Minhyuk Sung, and Sung-Eui Yoon.
\newblock Implicit lidar network: Lidar super-resolution via interpolation weight prediction.
\newblock In \emph{2022 International Conference on Robotics and Automation (ICRA)}, pages 8424--8430. IEEE, 2022{\natexlab{b}}.

\bibitem[Lang et~al.(2019)Lang, Vora, Caesar, Zhou, Yang, and Beijbom]{lang2019pointpillars}
Alex~H Lang, Sourabh Vora, Holger Caesar, Lubing Zhou, Jiong Yang, and Oscar Beijbom.
\newblock Pointpillars: Fast encoders for object detection from point clouds.
\newblock In \emph{Proceedings of the IEEE/CVF conference on computer vision and pattern recognition}, pages 12697--12705, 2019.

\bibitem[Lee et~al.(2021)Lee, Lee, and Song]{lee2021vision}
Seung~Hoon Lee, Seunghyun Lee, and Byung~Cheol Song.
\newblock Vision transformer for small-size datasets.
\newblock \emph{arXiv preprint arXiv:2112.13492}, 2021.

\bibitem[Li et~al.(2021)Li, Ismail, Shum, and Breckon]{li21durlar}
L. Li, K.N. Ismail, H.P.H. Shum, and T.P. Breckon.
\newblock Durlar: A high-fidelity 128-channel lidar dataset with panoramic ambient and reflectivity imagery for multi-modal autonomous driving applications.
\newblock In \emph{Proc. Int. Conf. on 3D Vision}. IEEE, 2021.

\bibitem[Liang et~al.(2021)Liang, Cao, Sun, Zhang, Van~Gool, and Timofte]{liang2021swinir}
Jingyun Liang, Jiezhang Cao, Guolei Sun, Kai Zhang, Luc Van~Gool, and Radu Timofte.
\newblock Swinir: Image restoration using swin transformer.
\newblock In \emph{Proceedings of the IEEE/CVF international conference on computer vision}, pages 1833--1844, 2021.

\bibitem[Lin et~al.(2022)Lin, Cheng, Zhong, Zhou, and Yang]{lin2022dynamic}
Yuankai Lin, Tao Cheng, Qi Zhong, Wending Zhou, and Hua Yang.
\newblock Dynamic spatial propagation network for depth completion.
\newblock In \emph{Proceedings of the AAAI Conference on Artificial Intelligence}, pages 1638--1646, 2022.

\bibitem[Ling et~al.(2023)Ling, Xing, Zhou, Cao, and Zhou]{ling2023panoswin}
Zhixin Ling, Zhen Xing, Xiangdong Zhou, Manliang Cao, and Guichun Zhou.
\newblock Panoswin: A pano-style swin transformer for panorama understanding.
\newblock In \emph{Proceedings of the IEEE/CVF Conference on Computer Vision and Pattern Recognition}, pages 17755--17764, 2023.

\bibitem[Liu et~al.(2021{\natexlab{a}})Liu, Sangineto, Bi, Sebe, Lepri, and De~Nadai]{liu2021efficient}
Yahui Liu, Enver Sangineto, Wei Bi, Nicu Sebe, Bruno Lepri, and Marco De~Nadai.
\newblock Efficient training of visual transformers with small datasets.
\newblock In \emph{Conference on Neural Information Processing Systems (NeurIPS)}, 2021{\natexlab{a}}.

\bibitem[Liu et~al.(2021{\natexlab{b}})Liu, Lin, Cao, Hu, Wei, Zhang, Lin, and Guo]{liu2021swintransformer}
Ze Liu, Yutong Lin, Yue Cao, Han Hu, Yixuan Wei, Zheng Zhang, Stephen Lin, and Baining Guo.
\newblock Swin transformer: Hierarchical vision transformer using shifted windows.
\newblock In \emph{Proceedings of the IEEE/CVF international conference on computer vision}, pages 10012--10022, 2021{\natexlab{b}}.

\bibitem[Loshchilov and Hutter(2017)]{adamw}
Ilya Loshchilov and Frank Hutter.
\newblock Decoupled weight decay regularization.
\newblock In \emph{International Conference on Learning Representations}, 2017.

\bibitem[Lu et~al.(2022)Lu, Li, Liu, Huang, Zhang, and Zeng]{lu2022transformerimagesr}
Zhisheng Lu, Juncheng Li, Hong Liu, Chaoyan Huang, Linlin Zhang, and Tieyong Zeng.
\newblock Transformer for single image super-resolution.
\newblock In \emph{Proceedings of the IEEE/CVF conference on computer vision and pattern recognition}, pages 457--466, 2022.

\bibitem[Park et~al.(2020)Park, Joo, Hu, Liu, and So~Kweon]{park2020non}
Jinsun Park, Kyungdon Joo, Zhe Hu, Chi-Kuei Liu, and In So~Kweon.
\newblock Non-local spatial propagation network for depth completion.
\newblock In \emph{Computer Vision--ECCV 2020: 16th European Conference, Glasgow, UK, August 23--28, 2020, Proceedings, Part XIII 16}, pages 120--136. Springer, 2020.

\bibitem[Qiu et~al.(2019)Qiu, Cui, Zhang, Zhang, Liu, Zeng, and Pollefeys]{qiu2019deeplidar}
Jiaxiong Qiu, Zhaopeng Cui, Yinda Zhang, Xingdi Zhang, Shuaicheng Liu, Bing Zeng, and Marc Pollefeys.
\newblock Deeplidar: Deep surface normal guided depth prediction for outdoor scene from sparse lidar data and single color image.
\newblock In \emph{Proceedings of the IEEE/CVF Conference on Computer Vision and Pattern Recognition}, pages 3313--3322, 2019.

\bibitem[Qiu et~al.(2022)Qiu, Anwar, and Barnes]{qiu2022pu}
Shi Qiu, Saeed Anwar, and Nick Barnes.
\newblock Pu-transformer: Point cloud upsampling transformer.
\newblock In \emph{Proceedings of the Asian Conference on Computer Vision}, pages 2475--2493, 2022.

\bibitem[Rho et~al.(2022)Rho, Ha, and Kim]{rho2022guideformer}
Kyeongha Rho, Jinsung Ha, and Youngjung Kim.
\newblock Guideformer: Transformers for image guided depth completion.
\newblock In \emph{Proceedings of the IEEE/CVF Conference on Computer Vision and Pattern Recognition}, pages 6250--6259, 2022.

\bibitem[Ronneberger et~al.(2015)Ronneberger, Fischer, and Brox]{ronneberger2015u}
Olaf Ronneberger, Philipp Fischer, and Thomas Brox.
\newblock U-net: Convolutional networks for biomedical image segmentation.
\newblock In \emph{Medical Image Computing and Computer-Assisted Intervention--MICCAI 2015: 18th International Conference, Munich, Germany, October 5-9, 2015, Proceedings, Part III 18}, pages 234--241. Springer, 2015.

\bibitem[Russakovsky et~al.(2015)Russakovsky, Deng, Su, Krause, Satheesh, Ma, Huang, Karpathy, Khosla, Bernstein, et~al.]{russakovsky2015imagenet}
Olga Russakovsky, Jia Deng, Hao Su, Jonathan Krause, Sanjeev Satheesh, Sean Ma, Zhiheng Huang, Andrej Karpathy, Aditya Khosla, Michael Bernstein, et~al.
\newblock Imagenet large scale visual recognition challenge.
\newblock \emph{International journal of computer vision}, 115:\penalty0 211--252, 2015.

\bibitem[Savkin et~al.(2022)Savkin, Wang, Wirkert, Navab, and Tombari]{savkin2022lidar}
Artem Savkin, Yida Wang, Sebastian Wirkert, Nassir Navab, and Federico Tombari.
\newblock Lidar upsampling with sliced wasserstein distance.
\newblock \emph{IEEE Robotics and Automation Letters}, 8\penalty0 (1):\penalty0 392--399, 2022.

\bibitem[Shan et~al.(2020)Shan, Wang, Chen, Szenher, and Englot]{shan2020simulation}
Tixiao Shan, Jinkun Wang, Fanfei Chen, Paul Szenher, and Brendan Englot.
\newblock Simulation-based lidar super-resolution for ground vehicles.
\newblock \emph{Robotics and Autonomous Systems}, 134:\penalty0 103647, 2020.

\bibitem[Shi et~al.(2016)Shi, Caballero, Husz{\'a}r, Totz, Aitken, Bishop, Rueckert, and Wang]{shi2016pixelshuffle}
Wenzhe Shi, Jose Caballero, Ferenc Husz{\'a}r, Johannes Totz, Andrew~P Aitken, Rob Bishop, Daniel Rueckert, and Zehan Wang.
\newblock Real-time single image and video super-resolution using an efficient sub-pixel convolutional neural network.
\newblock In \emph{Proceedings of the IEEE conference on computer vision and pattern recognition}, pages 1874--1883, 2016.

\bibitem[Sun et~al.(2023)Sun, Li, Zhang, Ma, Sheng, Cheng, Ma, Zhao, Zhang, Li, et~al.]{sun2023opdn}
Xiaopeng Sun, Weiqi Li, Zhenyu Zhang, Qiufang Ma, Xuhan Sheng, Ming Cheng, Haoyu Ma, Shijie Zhao, Jian Zhang, Junlin Li, et~al.
\newblock Opdn: Omnidirectional position-aware deformable network for omnidirectional image super-resolution.
\newblock In \emph{Proceedings of the IEEE/CVF Conference on Computer Vision and Pattern Recognition}, pages 1293--1301, 2023.

\bibitem[Tang et~al.(2020)Tang, Tian, Feng, Li, and Tan]{tang2020learning}
Jie Tang, Fei-Peng Tian, Wei Feng, Jian Li, and Ping Tan.
\newblock Learning guided convolutional network for depth completion.
\newblock \emph{IEEE Transactions on Image Processing}, 30:\penalty0 1116--1129, 2020.

\bibitem[Triess et~al.(2019)Triess, Peter, Rist, Enzweiler, and Z{\"o}llner]{triess2019lidarcnn}
Larissa~T Triess, David Peter, Christoph~B Rist, Markus Enzweiler, and J~Marius Z{\"o}llner.
\newblock Cnn-based synthesis of realistic high-resolution lidar data.
\newblock In \emph{2019 IEEE Intelligent Vehicles Symposium (IV)}, pages 1512--1519. IEEE, 2019.

\bibitem[Vaswani et~al.(2017)Vaswani, Shazeer, Parmar, Uszkoreit, Jones, Gomez, Kaiser, and Polosukhin]{vaswani2017attention}
Ashish Vaswani, Noam Shazeer, Niki Parmar, Jakob Uszkoreit, Llion Jones, Aidan~N Gomez, {\L}ukasz Kaiser, and Illia Polosukhin.
\newblock Attention is all you need.
\newblock \emph{Advances in neural information processing systems}, 30, 2017.

\bibitem[Wang et~al.(2023)Wang, Wang, She, Zhang, Qiu, and Xiao]{wang2023swintnfc}
Suhong Wang, Hongqing Wang, Shufeng She, Yanping Zhang, Qingju Qiu, and Zhifeng Xiao.
\newblock Swin-t-nfc crfs: An encoder--decoder neural model for high-precision uav positioning via point cloud super resolution and image semantic segmentation.
\newblock \emph{Computer Communications}, 197:\penalty0 52--60, 2023.

\bibitem[Wang et~al.(2022)Wang, Cun, Bao, Zhou, Liu, and Li]{wang2022uformer}
Zhendong Wang, Xiaodong Cun, Jianmin Bao, Wengang Zhou, Jianzhuang Liu, and Houqiang Li.
\newblock Uformer: A general u-shaped transformer for image restoration.
\newblock In \emph{Proceedings of the IEEE/CVF conference on computer vision and pattern recognition}, pages 17683--17693, 2022.

\bibitem[Wei and Zhang(2023)]{wei2023srno}
Min Wei and Xuesong Zhang.
\newblock Super-resolution neural operator.
\newblock In \emph{Proceedings of the IEEE/CVF Conference on Computer Vision and Pattern Recognition}, pages 18247--18256, 2023.

\bibitem[Wimmer et~al.(2022)Wimmer, Mehnert, and Condurache]{wimmer2022interspace}
Paul Wimmer, Jens Mehnert, and Alexandru Condurache.
\newblock Interspace pruning: Using adaptive filter representations to improve training of sparse cnns.
\newblock In \emph{Proceedings of the IEEE/CVF conference on computer vision and pattern recognition}, pages 12527--12537, 2022.

\bibitem[Yan et~al.(2021)Yan, Wang, Li, Zhang, Xu, Li, and Yang]{yan2107rignet}
Zhiqiang Yan, Kun Wang, Xiang Li, Zhenyu Zhang, Baobei Xu, Jun Li, and Jian Yang.
\newblock Rignet: Repetitive image guided network for depth completion.
\newblock In \emph{European Conference on Computer Vision}, 2021.

\bibitem[Yang et~al.(2019)Yang, Sun, Liu, Shen, and Jia]{yang2019std}
Zetong Yang, Yanan Sun, Shu Liu, Xiaoyong Shen, and Jiaya Jia.
\newblock Std: Sparse-to-dense 3d object detector for point cloud.
\newblock In \emph{Proceedings of the IEEE/CVF international conference on computer vision}, pages 1951--1960, 2019.

\bibitem[Ye et~al.(2021)Ye, Chen, Han, Wan, and Liao]{ye2021meta}
Shuquan Ye, Dongdong Chen, Songfang Han, Ziyu Wan, and Jing Liao.
\newblock Meta-pu: An arbitrary-scale upsampling network for point cloud.
\newblock \emph{IEEE transactions on visualization and computer graphics}, 28\penalty0 (9):\penalty0 3206--3218, 2021.

\bibitem[Yu et~al.(2023)Yu, Wang, Cao, Li, Shan, and Dong]{yu2023osrt}
Fanghua Yu, Xintao Wang, Mingdeng Cao, Gen Li, Ying Shan, and Chao Dong.
\newblock Osrt: Omnidirectional image super-resolution with distortion-aware transformer.
\newblock In \emph{Proceedings of the IEEE/CVF Conference on Computer Vision and Pattern Recognition}, pages 13283--13292, 2023.

\bibitem[Yu et~al.(2018)Yu, Li, Fu, Cohen-Or, and Heng]{yu2018pu}
Lequan Yu, Xianzhi Li, Chi-Wing Fu, Daniel Cohen-Or, and Pheng-Ann Heng.
\newblock Pu-net: Point cloud upsampling network.
\newblock In \emph{Proceedings of the IEEE conference on computer vision and pattern recognition}, pages 2790--2799, 2018.

\bibitem[Yun et~al.(2022)Yun, Lee, and Kim]{yun2022panovit}
Heeseung Yun, Sehun Lee, and Gunhee Kim.
\newblock Panoramic vision transformer for saliency detection in 360 videos.
\newblock In \emph{ECCV}, 2022.

\bibitem[Zhang et~al.(2019)Zhang, Jiang, Yang, Yamakawa, Shimada, and Kara]{zhang2019data}
Wentai Zhang, Haoliang Jiang, Zhangsihao Yang, Soji Yamakawa, Kenji Shimada, and Levent~Burak Kara.
\newblock Data-driven upsampling of point clouds.
\newblock \emph{Computer-Aided Design}, 112:\penalty0 1--13, 2019.

\bibitem[Zhang et~al.(2020)Zhang, Tian, Kong, Zhong, and Fu]{zhang2020rdnir}
Yulun Zhang, Yapeng Tian, Yu Kong, Bineng Zhong, and Yun Fu.
\newblock Residual dense network for image restoration.
\newblock \emph{TPAMI}, 2020.

\bibitem[Zhao et~al.(2022)Zhao, Liu, Zhong, Jiang, Gao, Li, and Ji]{zhao2022self}
Wenbo Zhao, Xianming Liu, Zhiwei Zhong, Junjun Jiang, Wei Gao, Ge Li, and Xiangyang Ji.
\newblock Self-supervised arbitrary-scale point clouds upsampling via implicit neural representation.
\newblock In \emph{Proceedings of the IEEE/CVF Conference on Computer Vision and Pattern Recognition}, pages 1999--2007, 2022.

\bibitem[Zhao et~al.(2021)Zhao, Hui, and Xie]{zhao2021sspu}
Yifan Zhao, Le Hui, and Jin Xie.
\newblock Sspu-net: Self-supervised point cloud upsampling via differentiable rendering.
\newblock In \emph{Proceedings of the 29th ACM International Conference on Multimedia}, pages 2214--2223, 2021.

\bibitem[Zov{\'a}thi et~al.(2023)Zov{\'a}thi, P{\'a}lffy, Jank{\'o}, and Benedek]{zovathi2023stdepthnet}
{\"O}rk{\'e}ny Zov{\'a}thi, Bal{\'a}zs P{\'a}lffy, Zsolt Jank{\'o}, and Csaba Benedek.
\newblock St-depthnet: A spatio-temporal deep network for depth completion using a single non-repetitive circular scanning lidar.
\newblock \emph{IEEE Robotics and Automation Letters}, 2023.

\end{thebibliography}
}

\clearpage
\setcounter{page}{1}
\maketitlesupplementary

\section{Network Details}
\label{sec:supp_network_details}
\begin{figure}[h!]
  \vspace{-2mm}
  \centering
    \includegraphics[width=0.45\textwidth]{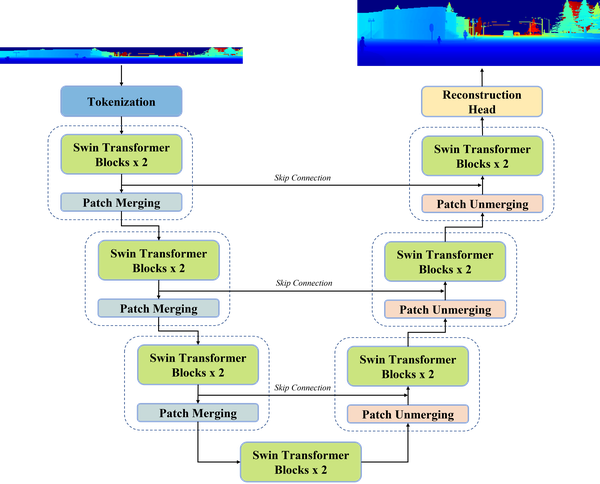}
    \vspace{-1mm}
\caption{\textbf{Network Architecture of \coolname{}:} The network has a symmetric design in the encoder and decoder. Each layer in the encoder consists of two Swin Transformer blocks and a patch merging layer which downscales the spatial resolution of feature maps. For the decoder, a patch unmerging layer succeeds two subsequent blocks for upscaling. A bottleneck layer is applied between the encoder and decoder to enhance the high-level feature representation. The range image is pre-processed with logarithmic transformation before being fed into the network.}
  \label{fig:network_architecture}
  \vspace{-2mm}
\end{figure} 
\subsection{Swin Transformer Block}
\begin{figure}[h]
  \vspace{-2mm}
  \centering
    \includegraphics[width=0.3\textwidth]{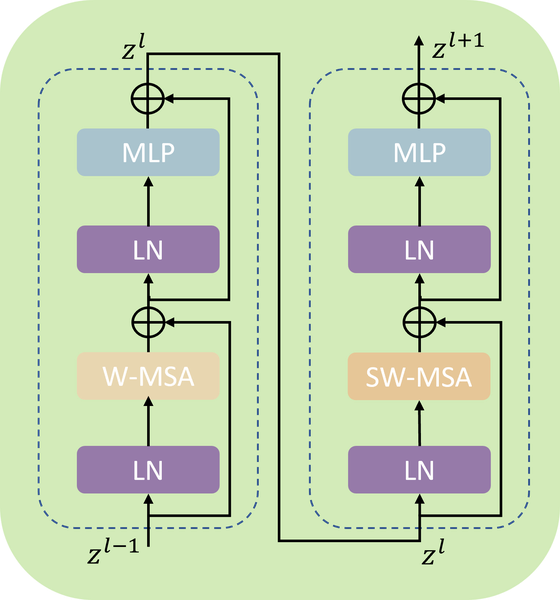}
    \vspace{-1mm}
\caption{Details of a Swin Transformer block. MLP: Multiple Layer Perceptron, LN: Layer Normalization, W-MSA: Window Multi-Head Self Attention, SW-MSA: Shifted Window Multi-Head Self Attention}
  \label{fig:swin_block}
  \vspace{-4mm}
\end{figure}

The structure of a Swin Transformer~\cite{liu2021swintransformer} block is presented in Fig.~\ref{fig:swin_block}. The block consists of two parts. The first part takes an input tensor with dimensions $H\times W\times C$ and initially, it performs layer normalization (LN) on the input. Then, it reshapes the feature vector into a tensor with dimensions $\frac{HW}{M^2} \times M^2 \times C$. This is achieved by partitioning the input into non-overlapping local windows of size $M \times M$, resulting in a total of $\frac{HW}{M^2}$ windows. For each of these local windows, the layer computes the query (Q), key (K), and value (V) matrices and applies the standard self-attention mechanism. The mathematical formulation is shown in Eq.~\ref{eq:self_attention}, where $B$ is a learnable relative positional encoding and $\sqrt{d}$ is a scaling factor. The second part has the same design and computes the attention with shifted windows. 

\begin{equation}
\label{eq:self_attention}
Attention(Q,K,V) = SoftMax(\frac{QK^T}{\sqrt{d}} + B)V , 
\end{equation}

\subsection{Monte Carlo Dropout}
\label{appendix:mcdrop}
\begin{figure}[h]
  \vspace{-2mm}
  \centering
    \begin{subfigure}[b]{0.3\textwidth}
    \includegraphics[width=\textwidth]{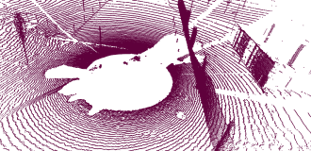}
    \caption{Ground-Truth}
    \end{subfigure}
    \begin{subfigure}[b]{0.3\textwidth}
    \includegraphics[width=\textwidth]{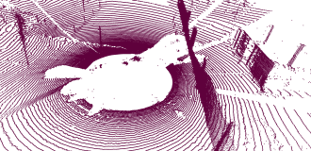}
    \caption{w/ MC-Dropout}
    \end{subfigure}
    \begin{subfigure}[b]{0.3\textwidth}
    \includegraphics[width=\textwidth]
    {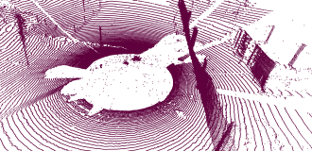}
    \caption{w/o MC-Dropout}
    \end{subfigure}
\caption{MC-Dropout cleans the range image by removing the noisy points.}
  \label{fig:monte_carlo_dropout}
  \vspace{-4mm}
\end{figure}

Inspired by the prior work in \cite{shan2020simulation}, we apply Monte Carlo Dropout (MC-Dropout)~\cite{gal2016dropout} to refine the prediction. For a given input $x$, a neural network with dropout makes a prediction $\hat{y}$ on each forward pass. Due to the active dropout layer in the inference, each forward pass effectively uses a differently configured model. By repeating the process $N$ times ($N = 50$ in this work), it yields a set of different outputs $\{\hat{y}_1, \hat{y}_2, ..., \hat{y}_N\}$. We then compute the mean of these $N$ outputs as the prediction ($\bar{y}$) and variance ($\bar{\sigma}$) which indicates the uncertainty. The formulation is shown in Eq.~\ref{eq:mc_dropout}.
\begin{equation}
\label{eq:mc_dropout}
\bar{y} = \frac{1}{N} \sum_{i=1}^{N} \hat{y}_i  \quad
\bar{\sigma}^2 = \frac{1}{N} \sum_{i=1}^{N} (\hat{y}_i - \bar{y})^2
\end{equation}
In terms of LiDAR point cloud, the uncertainty can be treated as the noise in the estimation of 3D coordinates. Hence, with a pre-defined threshold parameter $\lambda$, we can remove the noisy points and obtain the final prediction ($\bar{y*}$) with the decision rule in Eq.~\ref{eq:mc_dropout_decision_rule}. To obtain the final results, we chose a value of 0.03 for KITTI~\cite{geiger2012KITTI} and CARLA~\cite{kwon2022implicit}, and 0.0005 for DurLAR~\cite{li21durlar} dataset. 
\begin{equation}
\label{eq:mc_dropout_decision_rule}
    \bar{y*}= 
\begin{cases}
    \bar{y},& \text{if } \bar{\sigma} < \lambda * \bar{y}\\
    0,      & \text{otherwise}
\end{cases}
\end{equation}
To qualitatively show the effectiveness of post-processing the output, we visualize the range image in cases with and without MC-Dropout in Fig.~\ref{fig:monte_carlo_dropout}. 

\section{Additional Results}

\subsection{Discussion}
\begin{figure}[t]
  \vspace{-2mm}
  \centering
    \begin{subfigure}[b]{0.4\textwidth}
    \includegraphics[width=\textwidth]{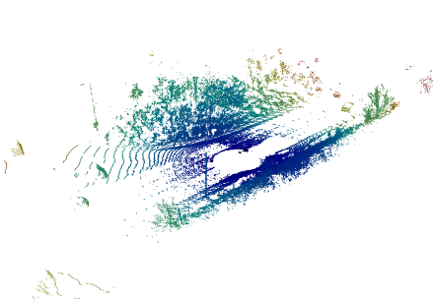}
    \caption{DurLAR}
    \end{subfigure}
    \begin{subfigure}[b]{0.4\textwidth}
    \includegraphics[width=\textwidth]{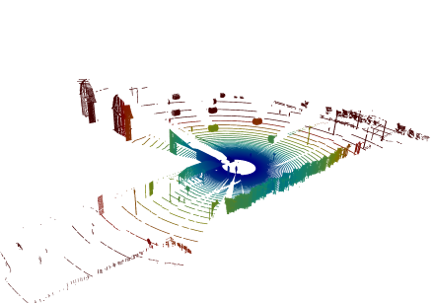}
    \caption{CARLA}
    \end{subfigure}
    \begin{subfigure}[b]{0.4\textwidth}
    \includegraphics[width=\textwidth]{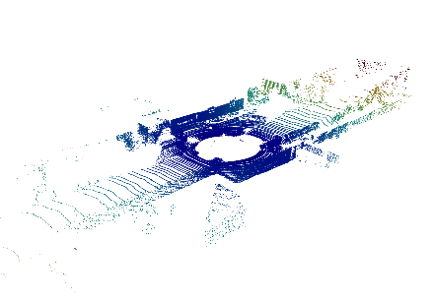}
    \caption{KITTI}
    \end{subfigure}
\caption{Example of LiDAR point cloud (ground-truth) from the test split of each dataset. DurLAR data contains more noise in the scan pattern and CARLA data is collected in a noiseless simulation environment.}
  \label{fig:example_data}
  \vspace{-4mm}
\end{figure} 
In the main experiments, our method presents a clear improvement in all three benchmarks. However, the enhancement for CARLA~\cite{kwon2022implicit} and DurLAR~\cite{li21durlar} is not as significant as for KITTI~\cite{geiger2012KITTI} dataset. In this section, we want to elaborate on this phenomenon.
\PAR{DurLAR:} The dataset contains 5 sub-datasets recorded from 5 different routes. To fairly create the train and test split from the dataset, we selected the "highway" route as the test sequence. This route contains lots of irregular objects such as trees and grass (as shown in Fig.~\ref{fig:example_data}), which are harder to upsample than regular objects such as houses or cars that are more frequently found in KITTI. Due to a sensor with a longer range, there are also roughly 10 times more points beyond 30 meters compared to KITTI. The point sparsity at higher ranges makes the reconstruction difficult. To support the argument, we additionally evaluated KITTI and DurLAR on points that are closer than 30 meters from the sensor origin in Tab.~\ref{tab:exp_within_30m}. The quantitative results indicate a similar trend of increase in Mean Absolute Error (MAE) and Chamfer Distance (CD) for the two datasets. %

\PAR{CARLA:} The simulated point clouds in \textbf{CARLA} exhibit a high level of orderliness. This benefits the interpolation methods (e.g. ILN) that are capable of upsampling while keeping more geometrical accuracy, however, the method suffers from repeatedly upsampling points in sparse regions. Conversely, our method distributes point upsampling across the scene more uniformly. We present visualizations for CARLA in columns I-J of Fig.~\ref{fig:vis_carla_durlar}. Quantitatively, in the main expeirments of LiDAR upsampling, the competitive score in IoU but remarkably lower score in Chamfer Distance (CD) for ILN also confirms the notion that our method is superior in uniformity of upsampling, resulting in lower error in Euclidean distance. 
\begin{table*}[h!]
\centering
\renewcommand \arraystretch{1.1}
\begin{tabular*}{\linewidth}{@{\extracolsep{\fill}} l c c c | c c c }
 & \multicolumn{3}{c}{KITTI [0-30m]} & \multicolumn{3}{c}{DurLAR [0-30m]}\\
\hline
Model & MAE $\downarrow$ & IoU $\uparrow$ & CD $\downarrow$ & MAE $\downarrow$ & IoU $\uparrow$ & CD $\downarrow$ \\
\hline
LIDAR-SR & 0.5393 & 0.1041 & 0.1042& 1.5180 & 0.1682 & 0.1079 \\
ILN &  0.9773 & 0.3409 & 0.1346& 1.5672 & 0.3418 & 0.0908\\
\coolname{} (Ours) & \underline{0.4174} & \underline{0.4462} & \underline{0.0286}& \underline{1.2408} & \underline{0.3662} & \underline{0.0223}\\
\coolname{}-L (Ours) & \textbf{0.3678} & \textbf{0.4673} & \textbf{0.0246} & \textbf{1.1645} & \textbf{0.3758} & \textbf{0.0208}\\
\arrayrulecolor{black}\hline 
\end{tabular*}
\captionof{table}{We evaluated upsampled point clouds within 30 meters. For Localization, we evaluated the performance of Range-MCL~\cite{chen2021range}, additionally on 64x1024 ground-truth (HR) and 16x1024 input (LR) of KITTI. *The model was evaluated on the full range as requested.}
\label{tab:exp_within_30m}
\end{table*}
\subsection{Inference Time}
We evaluate the inference time on the KITTI dataset~\cite{geiger2012KITTI}. A single forward pass for \coolname{}-L takes 1.33s. For the Monte-Carlo Dropout we run the separate forward passes in a batch and calculate the filtered result, which increases the effective inference time of \coolname{}-L to 1.61s (\coolname{} 1.13s). This is slower than ILN~\cite{kwon2022implicit} (1.14s) and LIDAR-SR~\cite{shan2020simulation} (1.16s) and comparable to Swin-IR~\cite{liang2021swinir} (1.63s).

\subsection{Model Parameters}
We present the number of model parameters in Tab.~\ref{tab:num_params}. Swin-IR~\cite{liang2021swinir} has fewer parameters because it utilizes a residual network design and hence, the scale of the decoder is much smaller than our method which deploys a U-Net-based architecture. ILN~\cite{kwon2022implicit} learns interpolation weights for neighboring points instead of upscaling the spatial resolution of features, so the network contains even fewer parameters at the cost of more memory for storing query points during training. LiDAR-SR~\cite{shan2020simulation} is comparable due to its similar network design as ours. It introduces a conventional CNN-based U-Net architecture for LiDAR upsampling and exceeds the total number of parameters in \coolname{}. In the main experiments, \coolname{}-L achieves more gain in reconstruction accuracy, however, the parameters are 4 times more than the baseline. A future study on shrinking network size is definitely of interest. Since we infer that most of the attention weights corresponding to those pixels with no occupancy (no beam return in 3D space) in the range image can be discarded without significantly compromising the accuracy, some model pruning techniques~\cite{wimmer2022interspace, kwon2022fast} can help to reduce the training costs, in addition, they can benefit shortening inference time as well.
\begin{table}[t]
\vspace{-2mm}
\renewcommand \arraystretch{0.9}
\resizebox{\columnwidth}{!}{%
    \begin{tabular}{ c|c|c|c|c } 
     Swin-IR  & LiDAR-SR & ILN* & \coolname{} (Ours)& \coolname{}-L (Ours)\\ 
    \hline
    11.8M & 34.6M&  1.3M & 27.1M & 108.1M\\ 
    \end{tabular}
    }
    \vspace{-4.7mm}
    \caption{Number of parameters of state-of-art methods in LiDAR upsampling and image super-resolution. We chose the network trained on KITTI dataset. }
    \label{tab:num_params}
\end{table}

\subsection{Generalization Capability}
We test the DurLAR dataset~\cite{li21durlar} using a model trained on the CARLA~\cite{kwon2022implicit} dataset to assess the generalization capability of different methods. The outcomes are displayed in Table~\ref{tab:generalization_ability}. Both sets of results, one from training on a different dataset and the other from training on the same dataset are presented to illustrate the degeneration in upsampling performance resulting from the domain shift. Although our method still outperforms most of the other methods in this case, except that it is slightly inferior to ILN~\cite{kwon2022implicit} in terms of IoU, we can observe significant drops in 3D evaluation metrics (IoU 17.3\% and Chamfer distance 97.2\%). 
\begin{table}[h]
\centering
\renewcommand \arraystretch{1.1}
\begin{tabular*}{\linewidth}{@{\extracolsep{\fill}} l c c c }
\hline
Model & MAE $\downarrow$ & IoU $\uparrow$ & CD $\downarrow$\\
\hline
SRNO~\cite{wei2023srno} & 2.5704  & 0.1467  & 0.9721 \\
HAT~\cite{chen2023hat}  & 2.6424  & 0.1561 & 0.8667 \\
SWIN-IR~\cite{liang2021swinir} &  2.2080   & 0.1809 & 0.4714 \\
LIIF~\cite{chen2021liif} & 1.9524  & 0.1757 & 0.1706 \\
\arrayrulecolor[rgb]{0.8,0.8,0.8} \hline
LIDAR-SR~\cite{shan2020simulation} &  2.0088 & 0.1352  & 0.4076 \\
ILN~\cite{kwon2022implicit} & 1.9044  & \textbf{0.3037} & 0.1291 \\
\coolname{} (Ours) & \textbf{1.8468} & 0.2995  & \textbf{0.1256} \\
\arrayrulecolor{black}\hline
\end{tabular*}
\caption{Quantitative comparison of the cross-dataset experiment: results are obtained by testing DurLAR dataset with the model trained on CARLA dataset. }
\label{tab:generalization_ability}
\end{table}

\begin{table}[b]
\centering
\renewcommand \arraystretch{1}
\scalebox{1}{\begin{tabular}{l c c c}
\hline
Model & MAE $\downarrow$ & IOU $\uparrow$ & CD $\downarrow$  \\
\hline
\multicolumn{4}{c}{Output Resolution: $64 \times 1024$} \\
\hline
*ILN~\cite{kwon2022implicit} & \textbf{1.4168} &  \textbf{0.3927} & \textbf{0.4447}\\
\coolname{} (Ours)  & 1.4776 &  0.3471 & 0.6087\\
\hline
\multicolumn{4}{c}{Input Resolution: $ 128 \times 2048$} \\
\hline
*ILN~\cite{kwon2022implicit} & \textbf{1.5368} & \textbf{0.3476} & 0.2993\\
\coolname{} (Ours)  & 1.5422 & 0.3451  & \textbf{0.2972}\\
\hline
\multicolumn{4}{c}{Output Resolution: $256 \times 4096$} \\
\hline
*ILN~\cite{kwon2022implicit} & 1.6088  & \textbf{0.2653} & 0.2219\\
\coolname{} (Ours)  & \textbf{1.5984} & 0.2523 & \textbf{0.1988}\\
\hline
\end{tabular}}
\caption{Quantitative results of scaling experiments. The input resolution is fixed with $16 \times 1024$ while the output resolution is varying. \textit{*We compare our method with Implicit LiDAR Network~\cite{kwon2022implicit} and obtain the results using the provided pretrained model.}}
\label{tab:different_scales}
\vspace{-1em}
\end{table}

\subsection{Different Upsampling Scales}
We test the adaptability of our approach in upsampling with different scaling ratios. As it is hard to find a real-world LiDAR configuration that can scan the same scene with multiple resolutions, we use CARLA~\cite{kwon2022implicit} dataset, created within the simulator and contains four different output resolutions: $16\times1024$, $64\times1024$, $128\times2048$ and $256\times4096$. We regard $16\times1024$ data as input and the other three as target resolutions for upsampling. In Tab.~\ref{tab:different_scales}, regarding to $\times 4$ upsampling with CARLA dataset, unlike upsampling from 32 to 128 channels, Implicit LiDAR Network (ILN)~\cite{kwon2022implicit} surpasses ours with respect to all evaluation metrics. Such an interpolation-based method tends to work more effectively for LiDAR upsampling in a noise-free environment, given that the target resolution and upsampling ratio are not exceedingly high. Raising the target resolution to $128 \times 2048$, ours is compatible with ILN, and coming to $\times 64$ upsampling, ours leads ILN by certain margins except IoU. %
\subsection{Downstream Tasks}
\begin{table*}[t!]
\centering
\renewcommand \arraystretch{0.9}
\begin{tabular*}{\linewidth}{@{\extracolsep{\fill}} l| c c c | c c }
\hline
& \multicolumn{3}{c}{Object Detection} & \multicolumn{2}{c}{Localization}\\
\hline
Model & Easy  & Moderate & Hard  & Location RMSE[m]$\downarrow$ & Yaw RMSE[deg]$\downarrow$\\
\hline
Low-Resolution*& 10.05	& 9.03 & 8.8 & 0.261 & 3.524 \\
LIDAR-SR~\cite{shan2020simulation} & 29.27	& 24.15 & 20.39 & 0.301 & 3.208 \\
ILN~\cite{kwon2022implicit} & 38.29 & 28.61 & 23.67& 0.263& 3.200\\
\coolname{} (Ours) & 50.23  & 37.57  & 32.12& 0.238 & 3.015 \\
\coolname{}-L (Ours) & 54.15 &	41.33  & 37.19& 0.238 & 2.981 \\
High-Resolution*& 73.33 & 62.78 & 59.63& 0.232 & 2.262\\
\hline
\hline
\end{tabular*}
\caption{We evaluated a pretrained Pointpillars model~\cite{lang2019pointpillars} and RangeMCL~\cite{chen2021range} model on $\times 4$ upsampled point clouds from the KITTI validation dataset~\cite{geiger2012KITTI} and report the overall results (averaged over classes 'Car', 'Cyclist' and 'Pedestrian' for object detection). Point clouds are generated by each model pretrained on KITTI Raw Dataset respectively. *Input and ground-truth data are tested as well.}
\label{tab:object_detection_and_localization}
\end{table*}
Apart from the main section regarding upsampling LiDAR point cloud and evaluating the results quantitatively and qualitatively with different metrics, we want to further assess how well the upsampling process preserves the shape of foreground objects, hence we conduct an experiment using an object detection model applied to the upsampled point clouds. In particular, we assessed our method on two downstream tasks: 3D object detection and localization, which can certainly benefit from the densification of the LiDAR point cloud. \\
For object detection, we refer to PointPillar~\cite{lang2019pointpillars}, which is one of the most commonly used models for 3D object detection. To make a fair comparison, we directly use a model pre-trained on KITTI Object Dataset~\cite{geiger2012KITTI} and generate the 3D bounding boxes on the generated, ground truth and low-resolution point clouds. For evaluation, we utilize the Average Precision (AP) with 11 precision-recall positions at an overlap threshold of 0.7 IoU and additionally calculate the mean Average Precision (mAP) over three different classes. Results are shown in Tab.~\ref{tab:object_detection_and_localization}. Compared to conventional range image upsampling techniques~\cite{shan2020simulation, kwon2022implicit}, our approach presents significantly superior results in 3D object detection. Although there remains a clear gap to the ground truth point cloud, the incorporation of an upsampling network to generate a denser point cloud proves beneficial in detecting more objects and achieving more accurate detection.\\
For localization, we chose RangeMCL~\cite{chen2021range}, which builds an observation model formulated from the discrepancies between real and rendered range images from a mesh map for a Monte Carlo Localization framework, to recalibrate the importance of weights attributed to each particle. We followed the evaluation steps introduced in the work and assessed the localization pipeline on the point clouds upsampled from different methods. In Tab.~\ref{tab:object_detection_and_localization}, it shows that upsampling the point cloud with our method generally improves the results of localization compared to using the low-resolution one directly while ILN~\cite{kwon2022implicit} and LiDAR-SR~\cite{shan2020simulation} lead to a larger error in location.
\section{Additional Qualitative Results}
Besides more results of KITTI dataset shown in Fig~\ref{fig:vis_kitti}, we provide additional visualization coming from the other two datasets in Fig~\ref{fig:vis_carla_durlar}. %

\ifdefined \FIGINSUPP
 \begin{figure*}[t]
  \hspace{-5mm}
    \centering
    \begin{tabular*}{\linewidth}{m{0.1cm}m{3cm}m{3cm}m{3cm}m{3cm}m{3cm}}
    \rotatebox{90}{\fontsize{7}{10}\selectfont Input}&\includegraphics[width=0.18\textwidth]{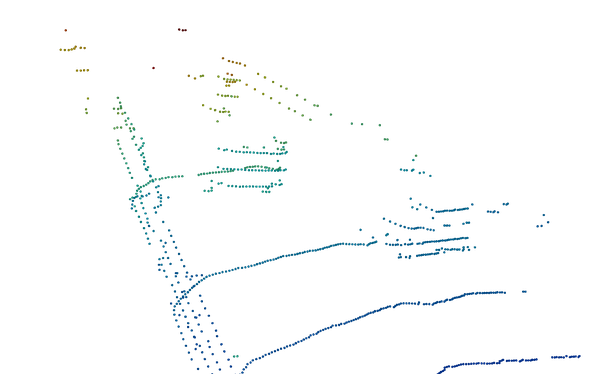}&\includegraphics[width=0.18\textwidth]{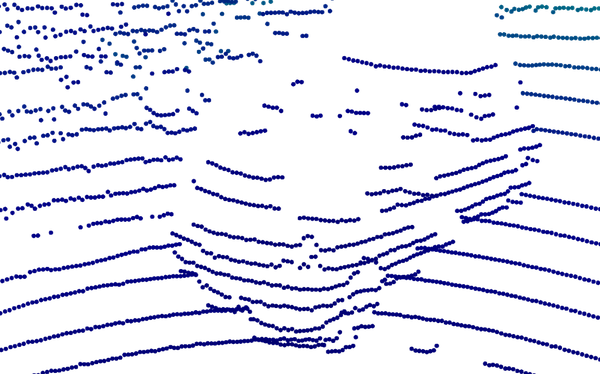}&\includegraphics[width=0.18\textwidth]{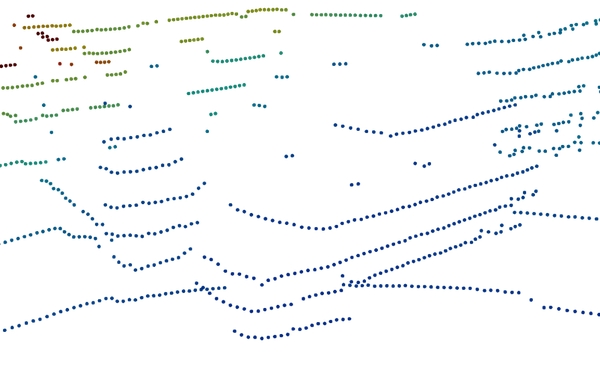}&\includegraphics[width=0.18\textwidth]{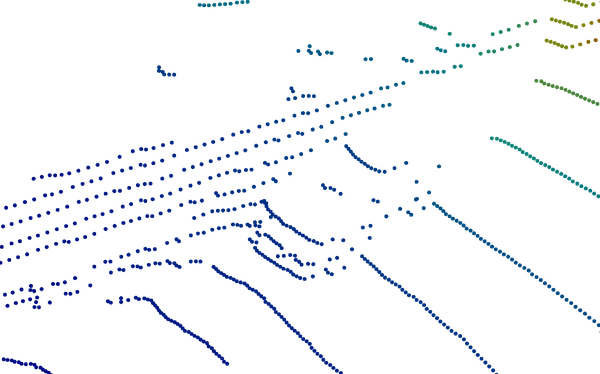}&\includegraphics[width=0.18\textwidth]{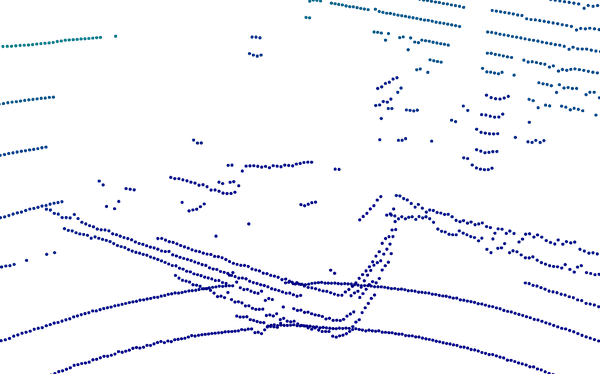}\\
    \rotatebox{90}{\fontsize{7}{10}\selectfont Bilinear} &\includegraphics[width=0.18\textwidth]{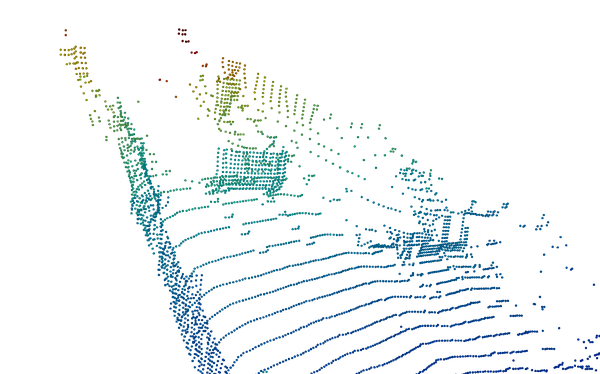}&\includegraphics[width=0.18\textwidth]{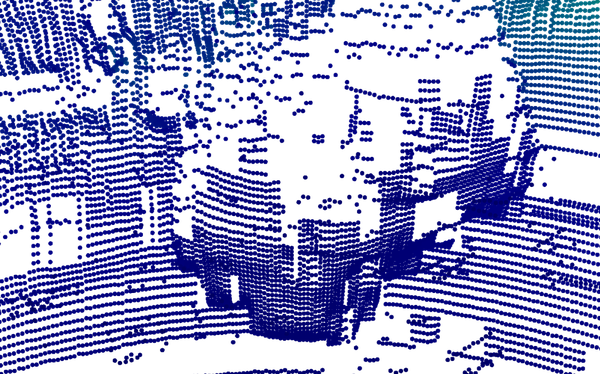}&\includegraphics[width=0.18\textwidth]{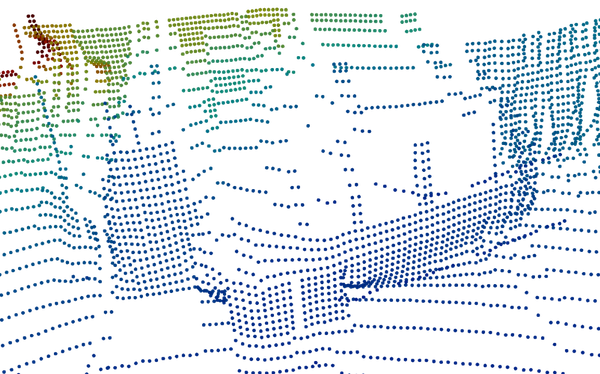}&\includegraphics[width=0.18\textwidth]{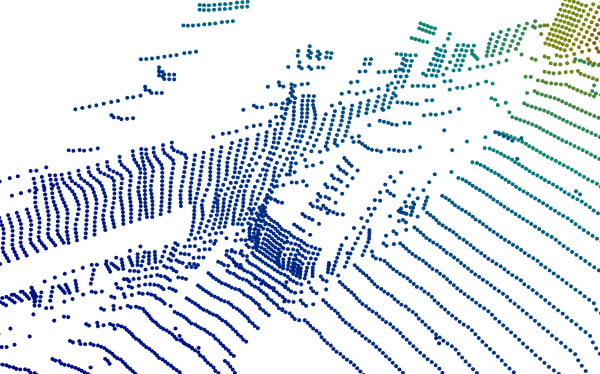}&\includegraphics[width=0.18\textwidth]{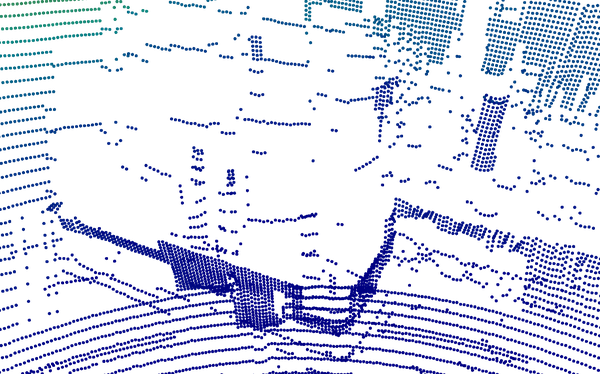}\\
    \rotatebox{90}{\fontsize{7}{10}\selectfont HAT~\cite{chen2023hat}} &\includegraphics[width=0.18\textwidth]{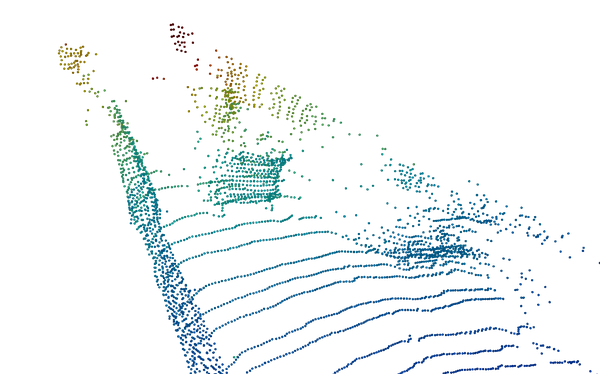}&\includegraphics[width=0.18\textwidth]{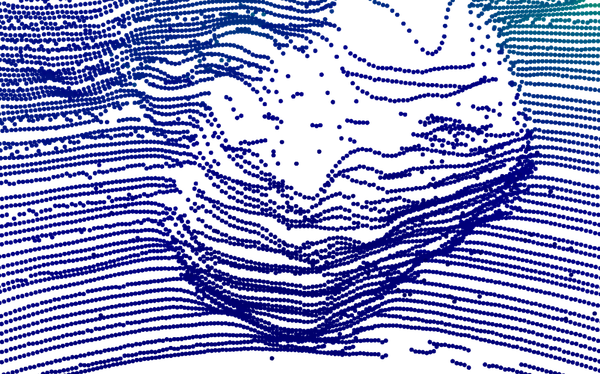}&\includegraphics[width=0.18\textwidth]{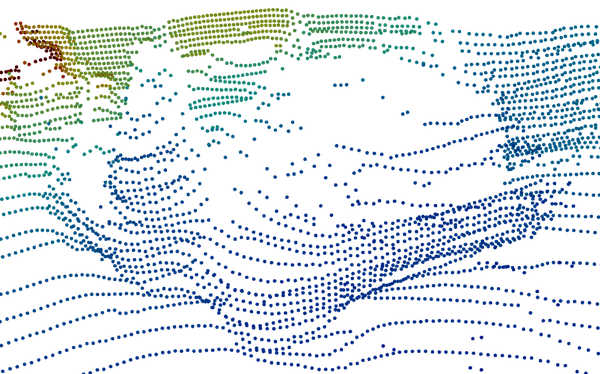}&\includegraphics[width=0.18\textwidth]{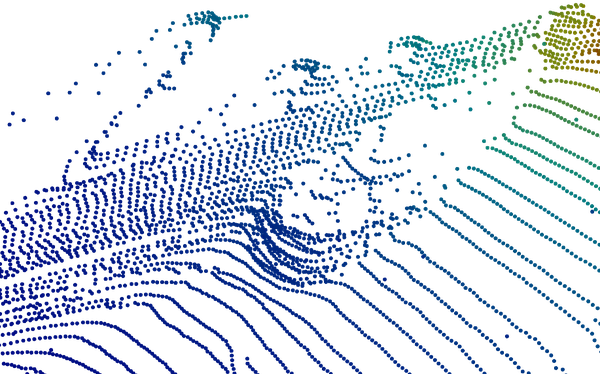}&\includegraphics[width=0.18\textwidth]{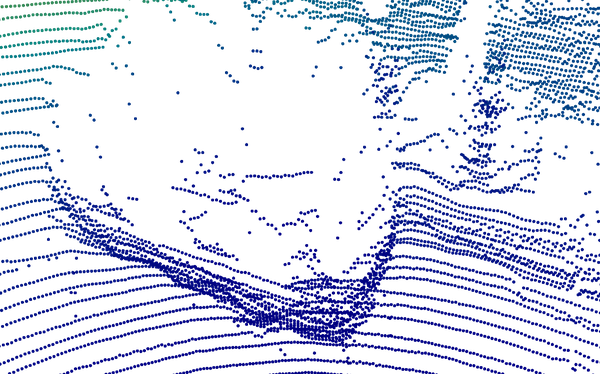}\\
    \rotatebox{90}{\fontsize{7}{10}\selectfont SRNO~\cite{wei2023srno}}&\includegraphics[width=0.18\textwidth]{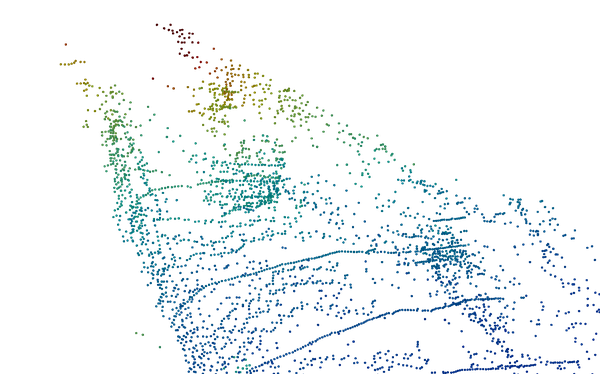}&\includegraphics[width=0.18\textwidth]{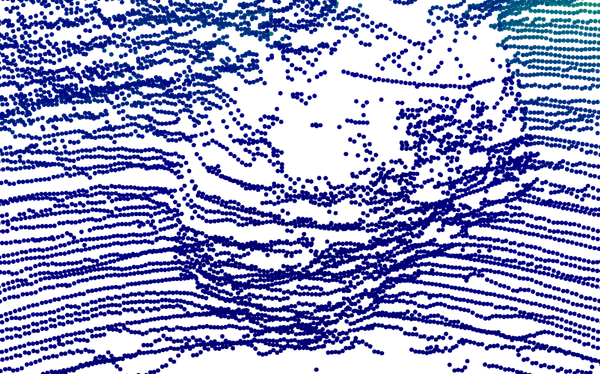}&\includegraphics[width=0.18\textwidth]{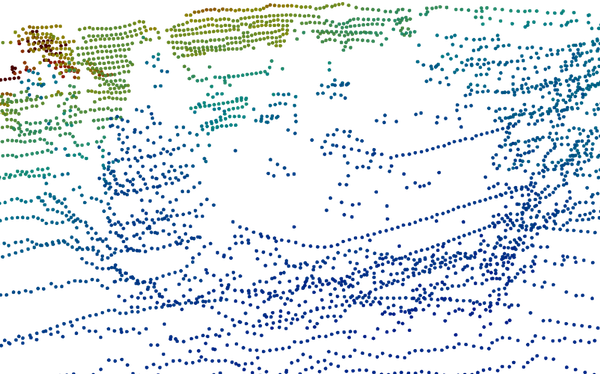}&\includegraphics[width=0.18\textwidth]{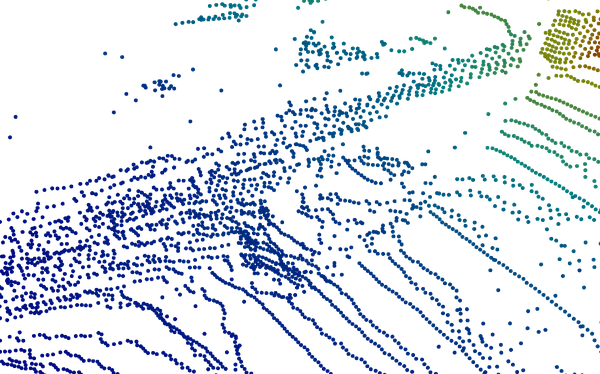}&\includegraphics[width=0.18\textwidth]{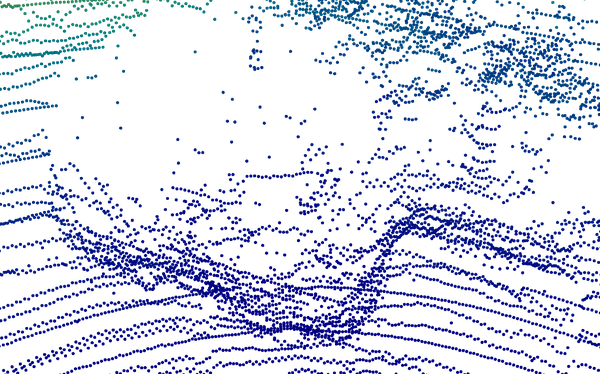}\\
    \rotatebox{90}{\fontsize{7}{10}\selectfont Swin-IR~\cite{liang2021swinir}}&\includegraphics[width=0.18\textwidth]{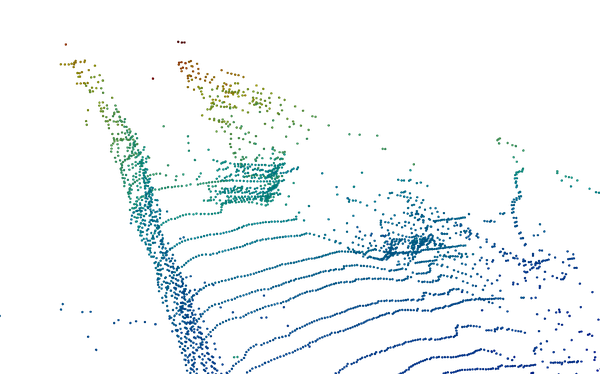}&\includegraphics[width=0.18\textwidth]{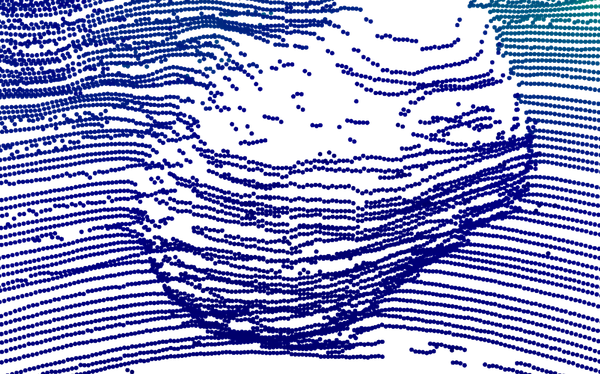}&\includegraphics[width=0.18\textwidth]{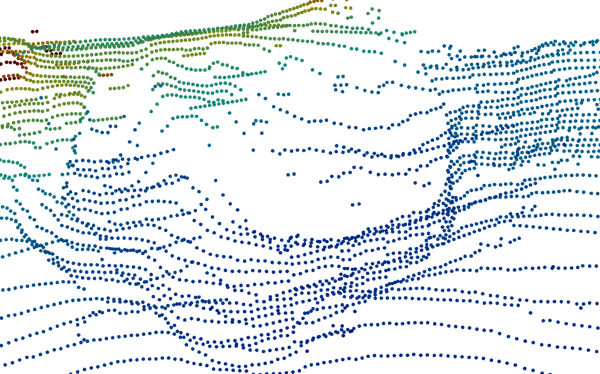}&\includegraphics[width=0.18\textwidth]{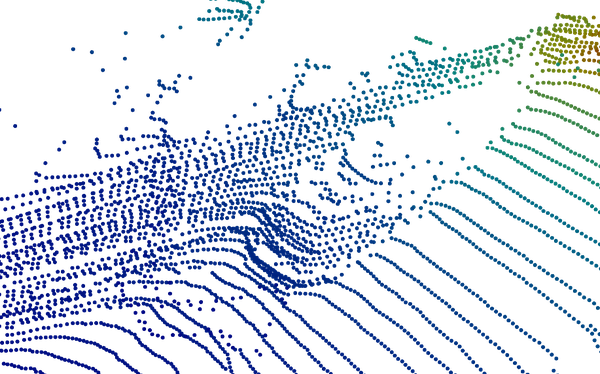}&\includegraphics[width=0.18\textwidth]{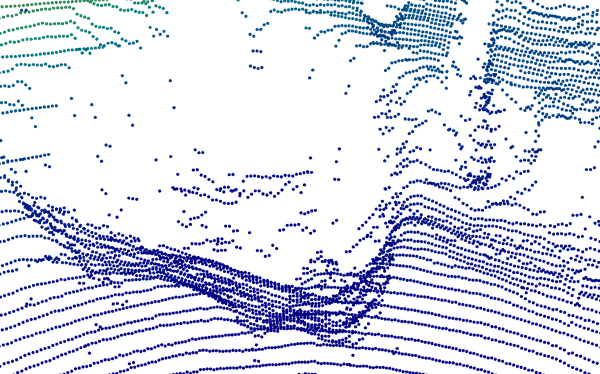}\\
    \rotatebox{90}{\fontsize{7}{10}\selectfont LIIF~\cite{chen2021liif}}&\includegraphics[width=0.18\textwidth]{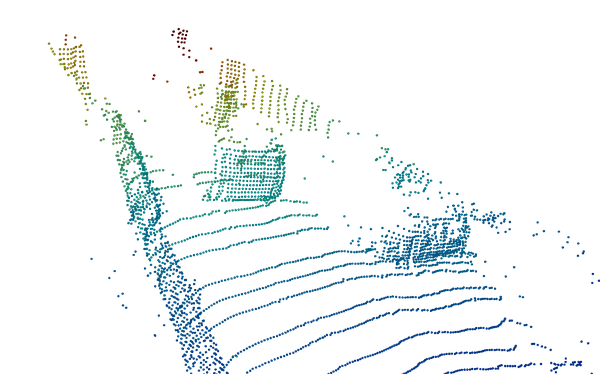}&\includegraphics[width=0.18\textwidth]{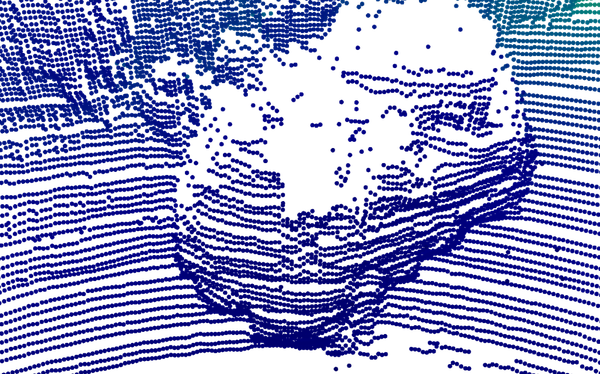}&\includegraphics[width=0.18\textwidth]{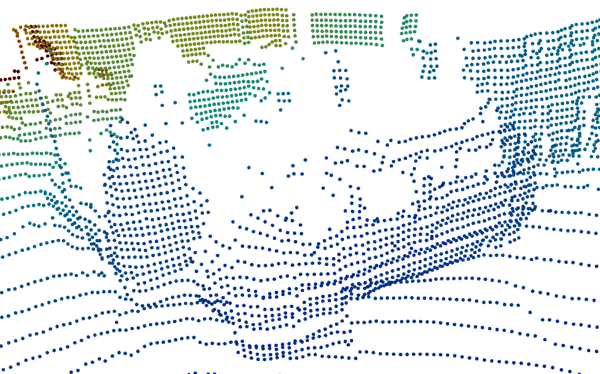}&\includegraphics[width=0.18\textwidth]{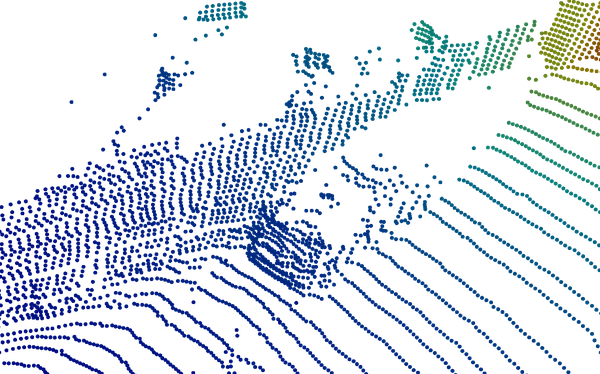}&\includegraphics[width=0.18\textwidth]{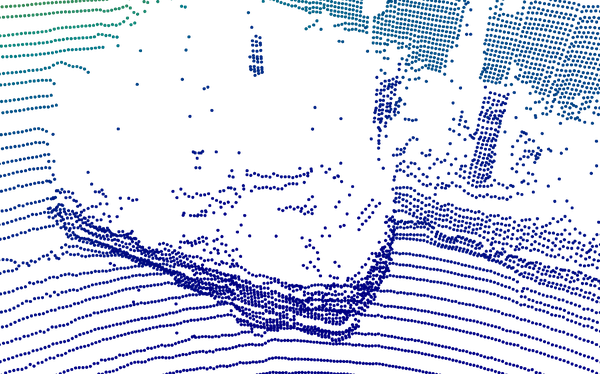}\\
    \rotatebox{90}{\fontsize{7}{10}\selectfont LiDA-SR~\cite{shan2020simulation}}&\includegraphics[width=0.18\textwidth]{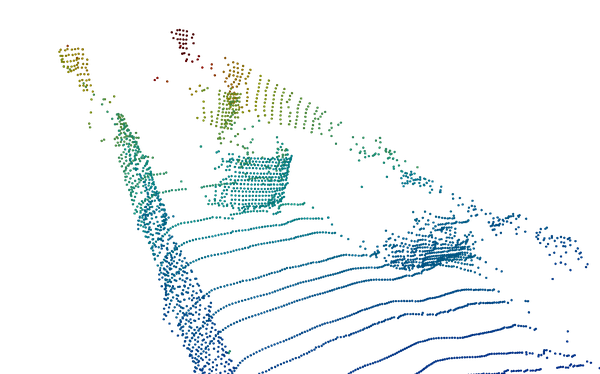}&\includegraphics[width=0.18\textwidth]{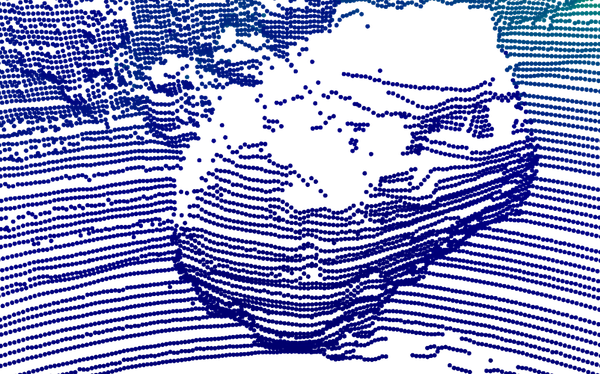}&\includegraphics[width=0.18\textwidth]{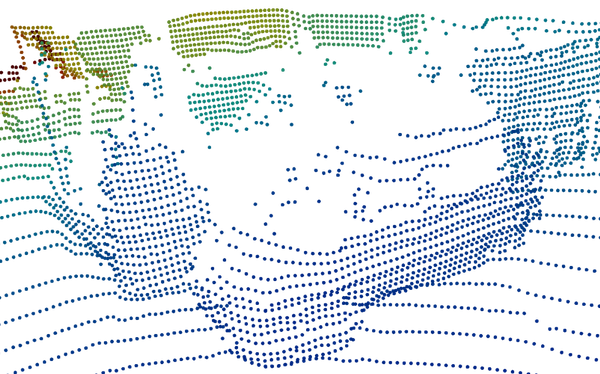}&\includegraphics[width=0.18\textwidth]{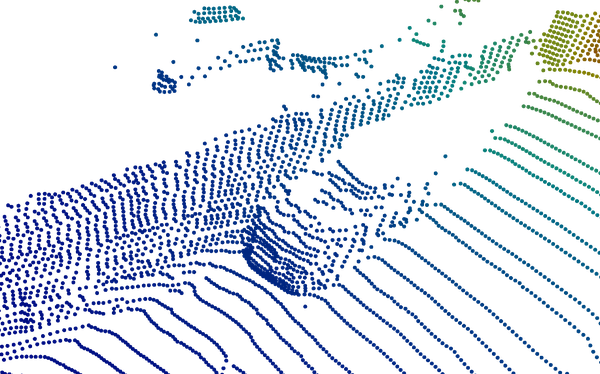}&\includegraphics[width=0.18\textwidth]{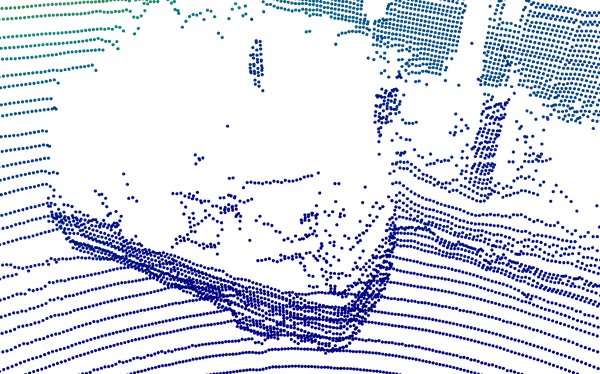}\\
    \rotatebox{90}{\fontsize{7}{10}\selectfont ILN~\cite{kwon2022implicit}}&\includegraphics[width=0.18\textwidth]{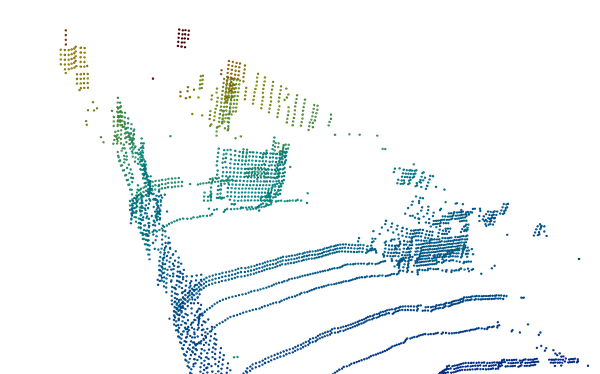}&\includegraphics[width=0.18\textwidth]{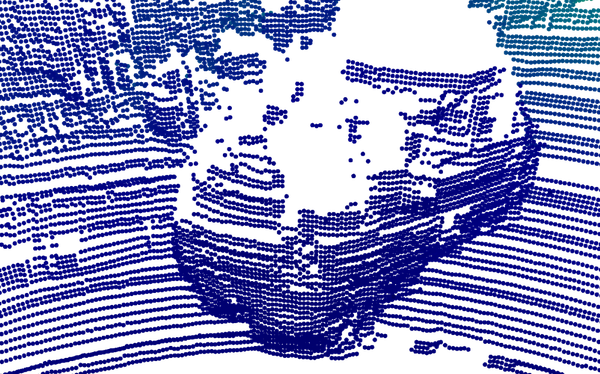}&\includegraphics[width=0.18\textwidth]{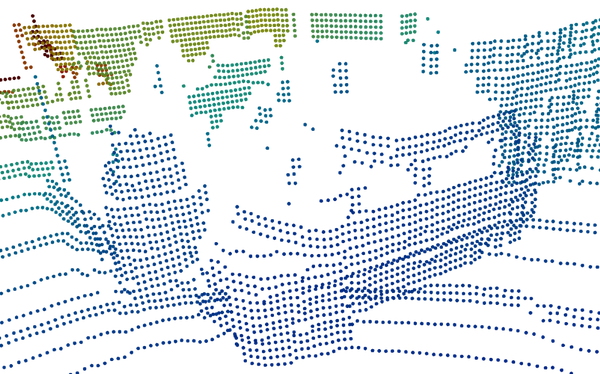}&\includegraphics[width=0.18\textwidth]{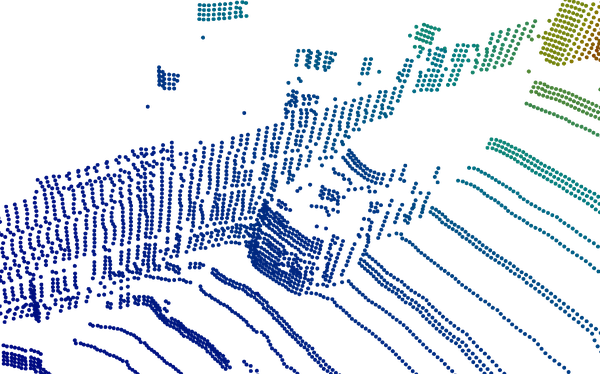}&\includegraphics[width=0.18\textwidth]{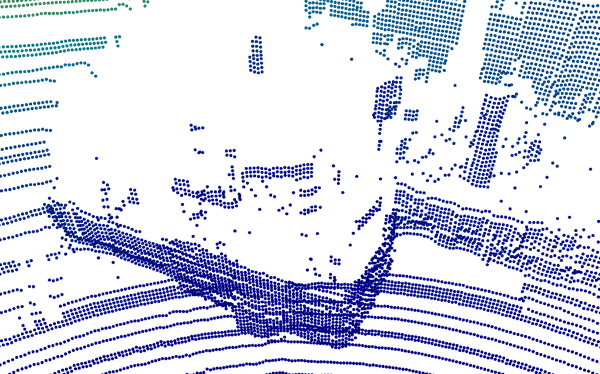}\\
    \rotatebox{90}{\fontsize{7}{10}\selectfont \coolname{}(Ours)}&
    \includegraphics[width=0.18\textwidth]{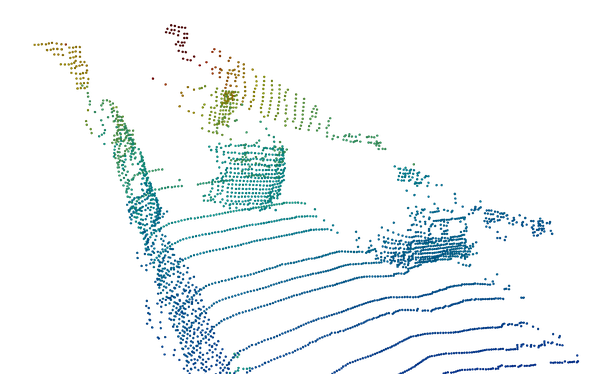}&\includegraphics[width=0.18\textwidth]{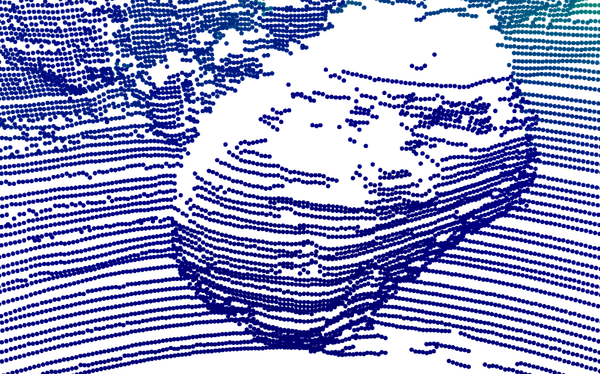}&\includegraphics[width=0.18\textwidth]{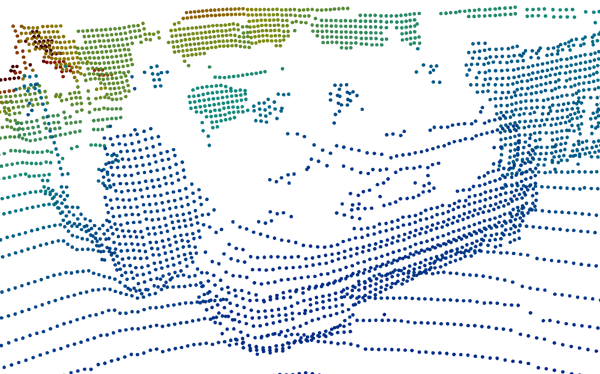}&\includegraphics[width=0.18\textwidth]{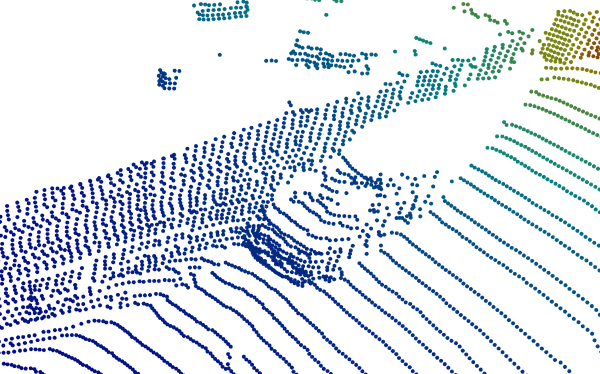}&\includegraphics[width=0.18\textwidth]{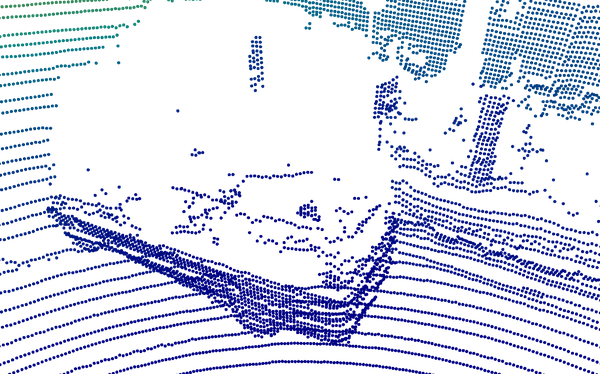}\\
    \rotatebox{90}{\fontsize{7}{10}\selectfont GT}&\includegraphics[width=0.18\textwidth]{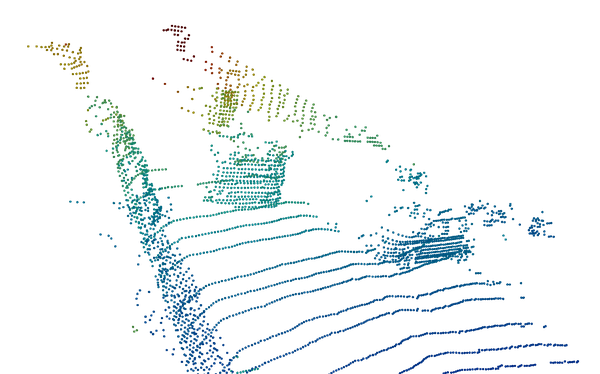}&\includegraphics[width=0.18\textwidth]{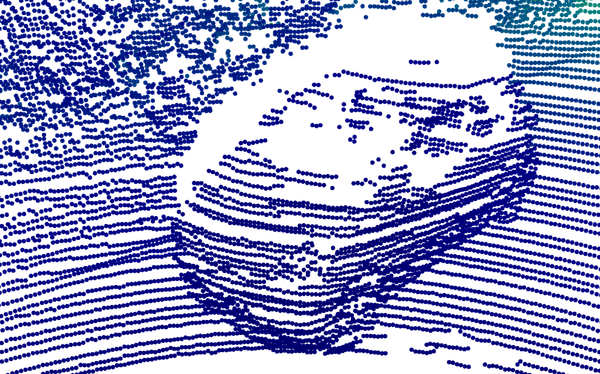}&\includegraphics[width=0.18\textwidth]{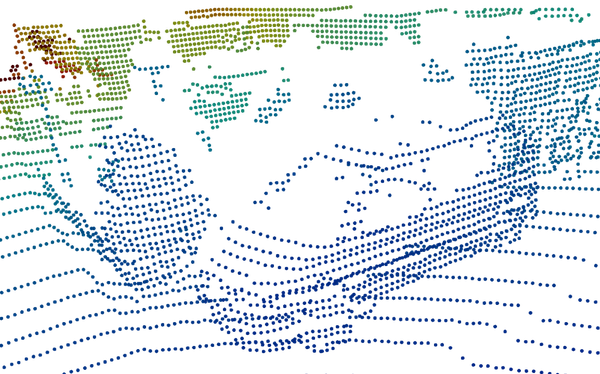}&\includegraphics[width=0.18\textwidth]{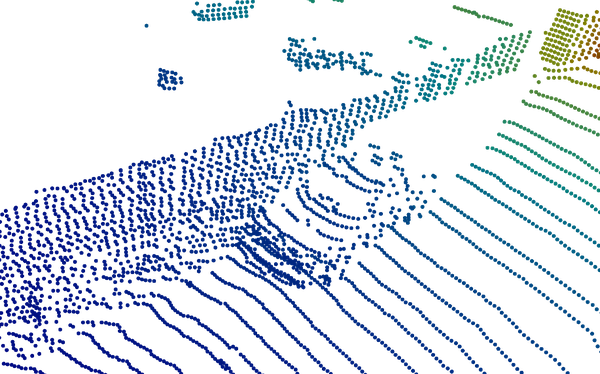}&\includegraphics[width=0.18\textwidth]{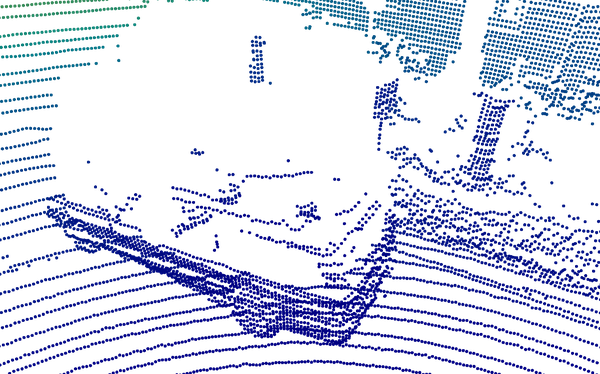}\\
    \quad & \hspace{13mm} A & \hspace{13mm} B & \hspace{13mm} C& \hspace{13mm} D& \hspace{13mm} E\\
    
    \end{tabular*}
    \caption{Additional Qualitative Results of KITTI Dataset with full comparison: Our approach outperforms all state-of-the-art approaches in upsampling the point cloud in a geometry-aware manner, specifically in terms of reconstruction of objects like cars, walls, and lidar sweeps while producing much fewer noisy points.}
    \label{fig:vis_kitti}
\end{figure*}

 \begin{figure*}[t]
 \vspace{-5mm}
  \hspace{-5mm}
    \centering
    \begin{tabular*}{\linewidth}{m{0.1cm}m{3cm}m{3cm}m{3cm}m{3cm}m{3cm}}
    \rotatebox{90}{\fontsize{7}{10}\selectfont Input}&\includegraphics[width=0.18\textwidth]{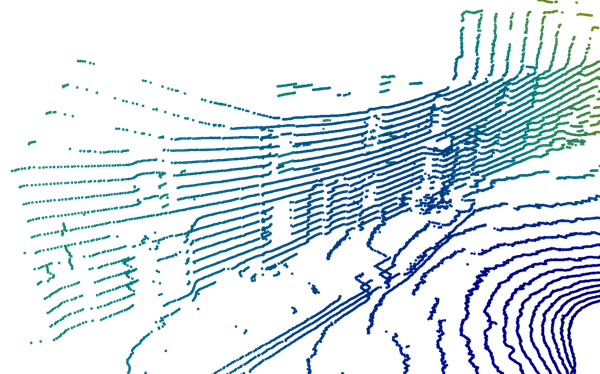}&\includegraphics[width=0.18\textwidth]{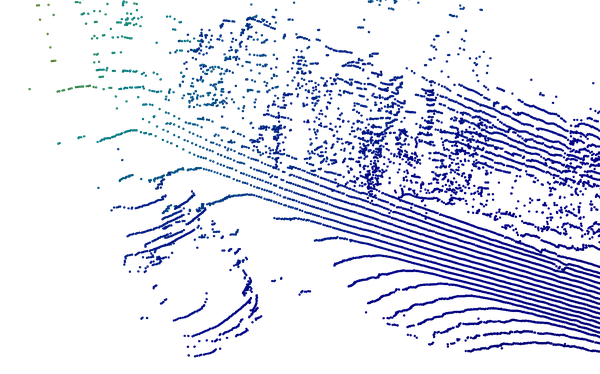}&\includegraphics[width=0.18\textwidth]{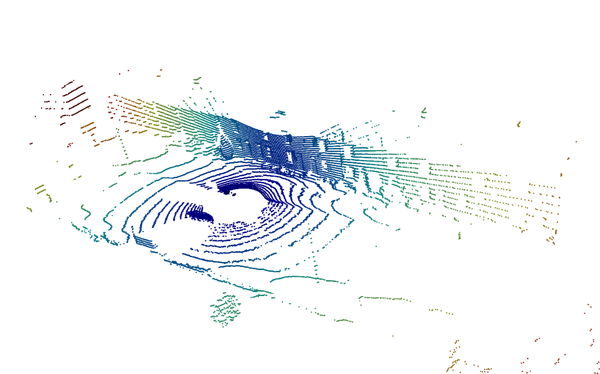}&\includegraphics[width=0.18\textwidth]{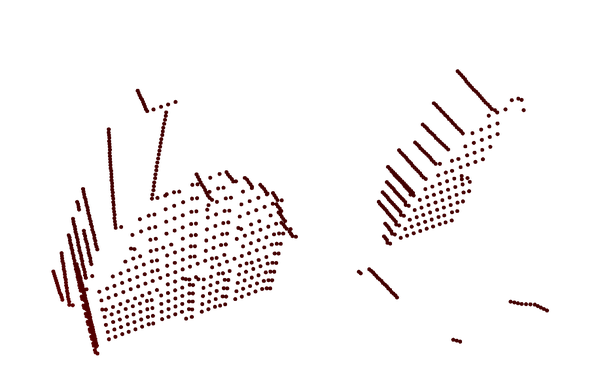}&\includegraphics[width=0.18\textwidth]{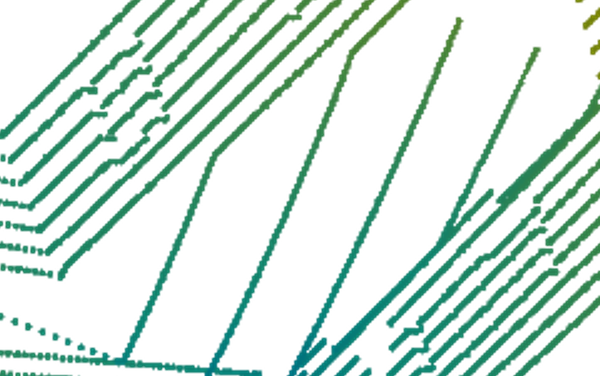}\\
    \rotatebox{90}{\fontsize{7}{10}\selectfont Bilinear} &\includegraphics[width=0.18\textwidth]{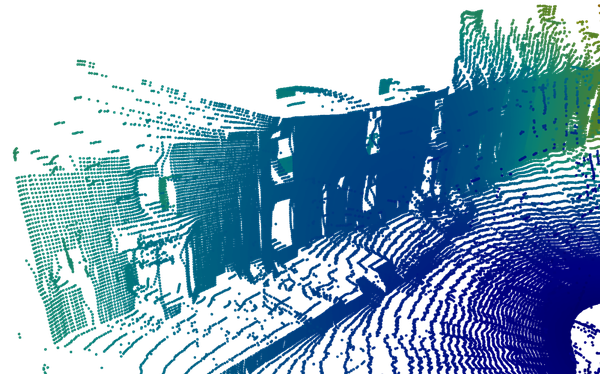}&\includegraphics[width=0.18\textwidth]{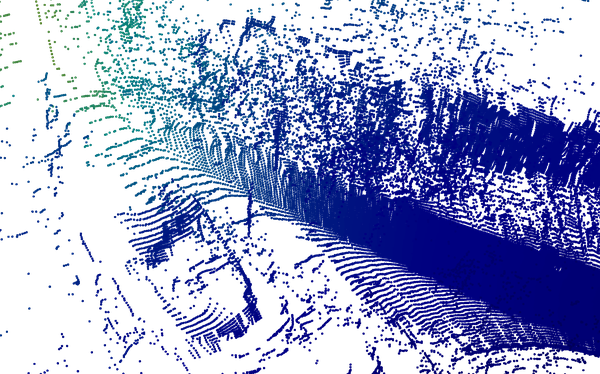}&\includegraphics[width=0.18\textwidth]{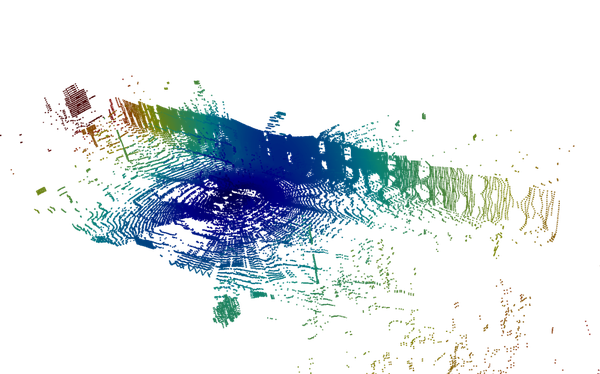}&\includegraphics[width=0.18\textwidth]{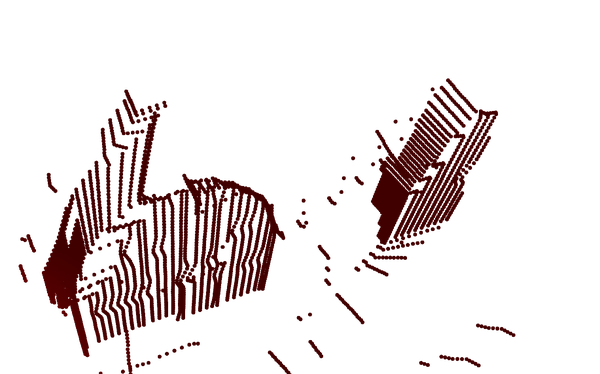}&\includegraphics[width=0.18\textwidth]{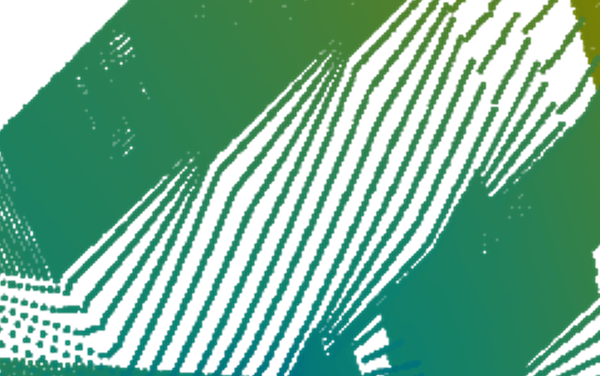}\\
    \rotatebox{90}{\fontsize{7}{10}\selectfont HAT~\cite{chen2023hat}} &\includegraphics[width=0.18\textwidth]{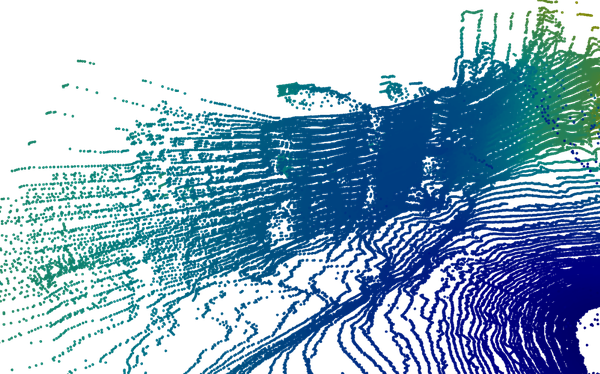}&\includegraphics[width=0.18\textwidth]{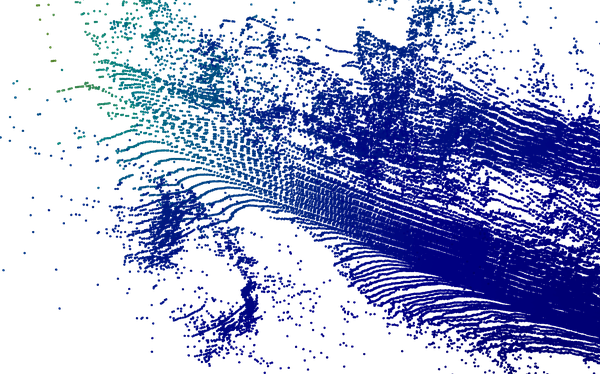}&\includegraphics[width=0.18\textwidth]{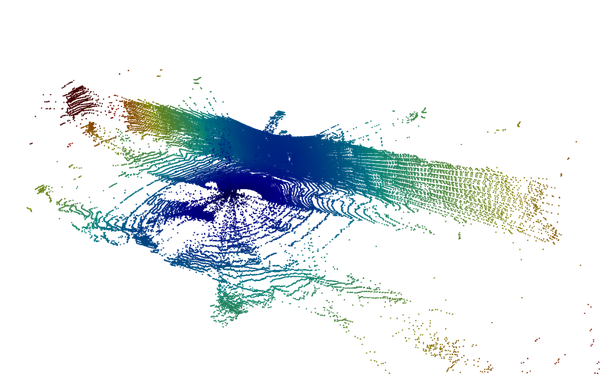}&\includegraphics[width=0.18\textwidth]{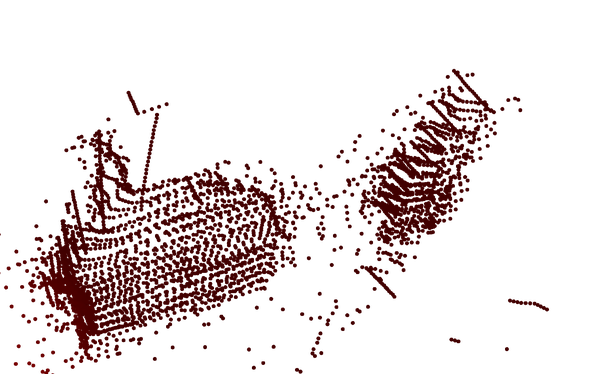}&\includegraphics[width=0.18\textwidth]{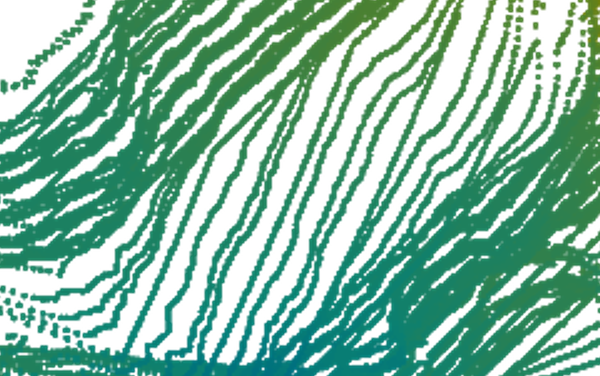}\\
    \rotatebox{90}{\fontsize{7}{10}\selectfont SRNO~\cite{wei2023srno}}&\includegraphics[width=0.18\textwidth]{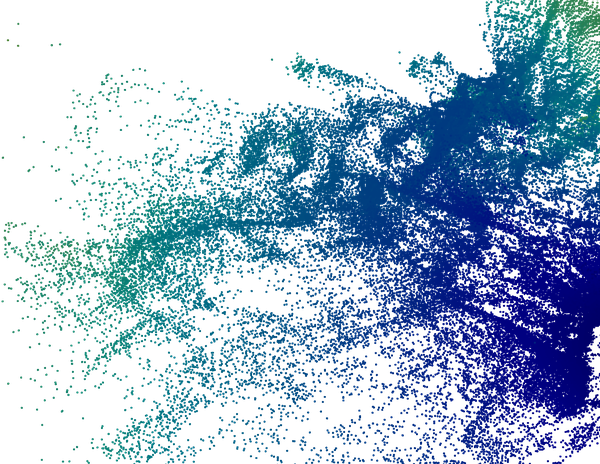}&\includegraphics[width=0.18\textwidth]{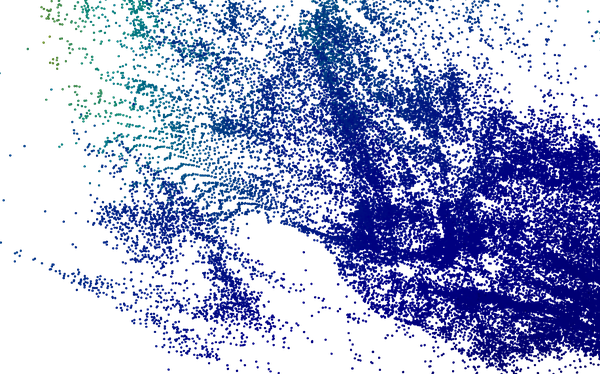}&\includegraphics[width=0.18\textwidth]{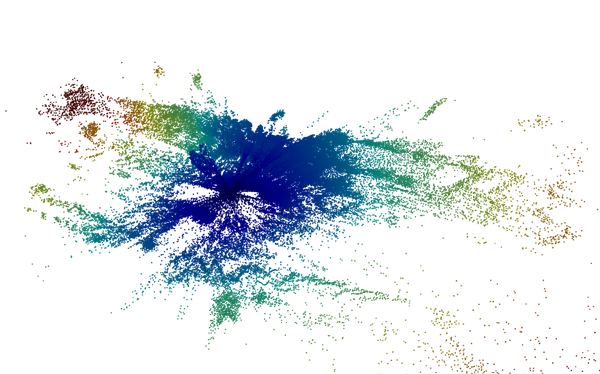}&\includegraphics[width=0.18\textwidth]{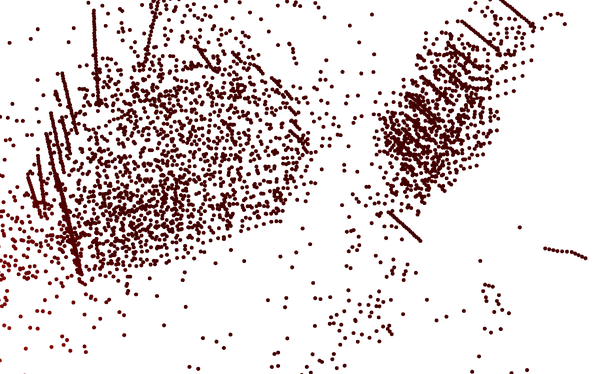}&\includegraphics[width=0.18\textwidth]{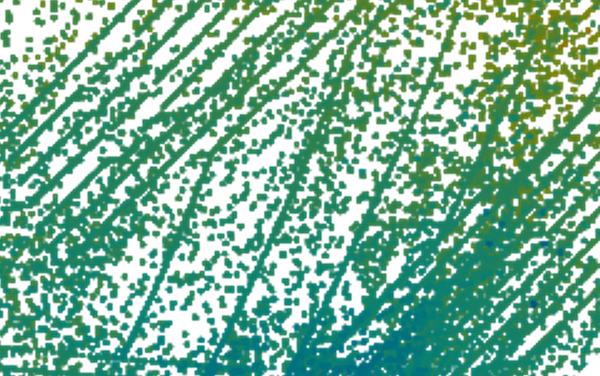}\\
    \rotatebox{90}{\fontsize{7}{10}\selectfont Swin-IR~\cite{liang2021swinir}}&\includegraphics[width=0.18\textwidth]{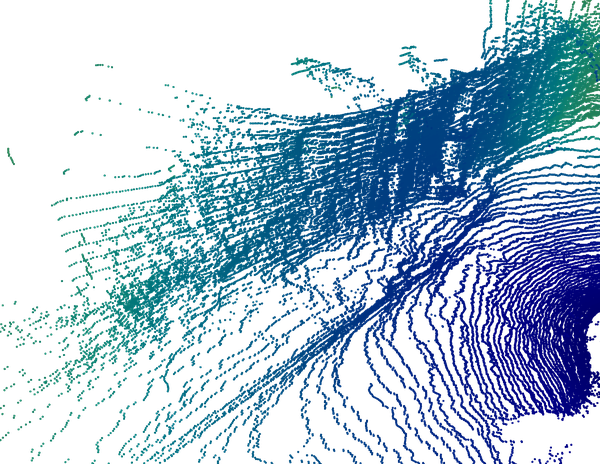}&\includegraphics[width=0.18\textwidth]{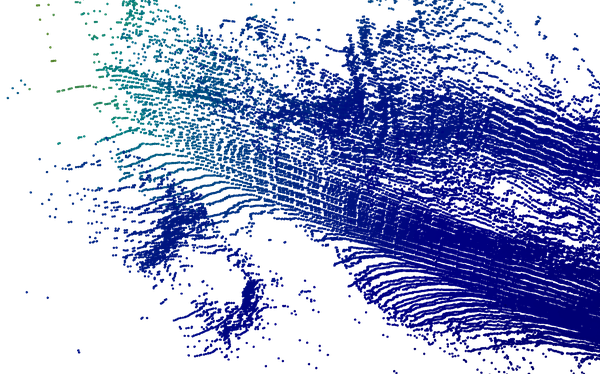}
    &\includegraphics[width=0.18\textwidth]{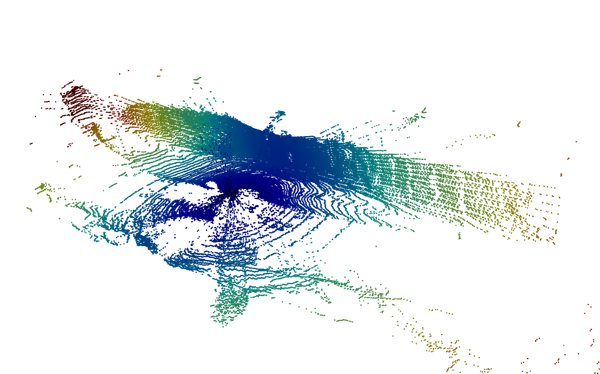}&\includegraphics[width=0.18\textwidth]{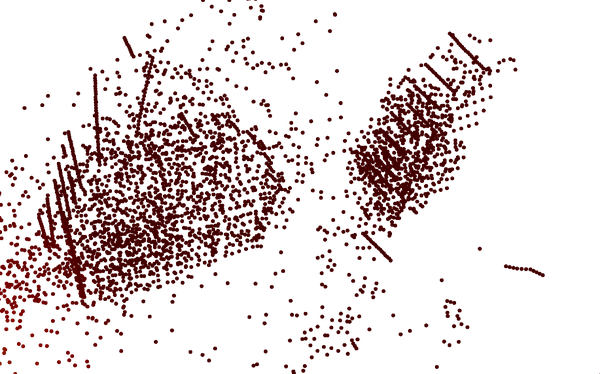}&\includegraphics[width=0.18\textwidth]{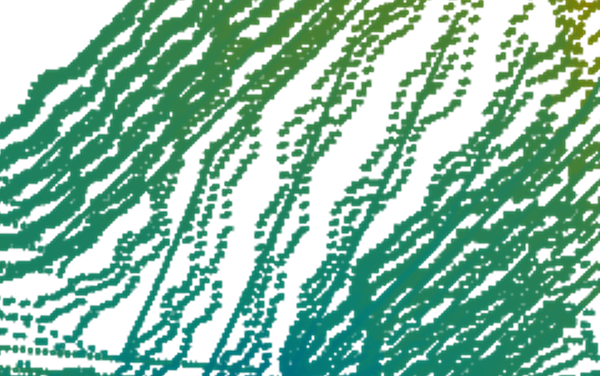}\\
    \rotatebox{90}{\fontsize{7}{10}\selectfont LIIF~\cite{chen2021liif}}&\includegraphics[width=0.18\textwidth]{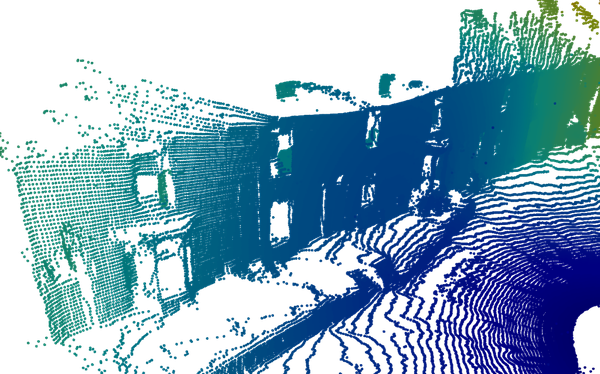}&\includegraphics[width=0.18\textwidth]{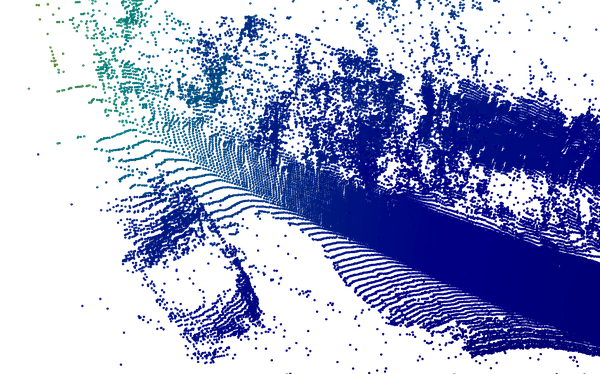}&\includegraphics[width=0.18\textwidth]{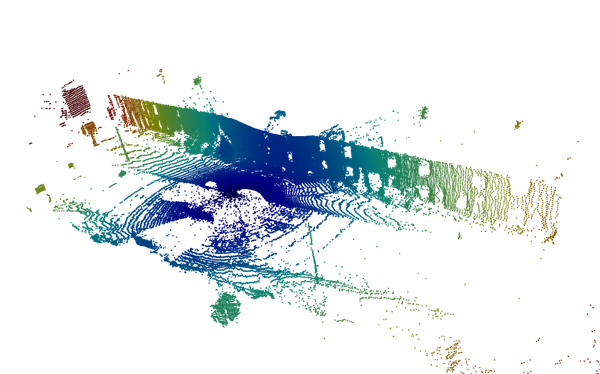}&\includegraphics[width=0.18\textwidth]{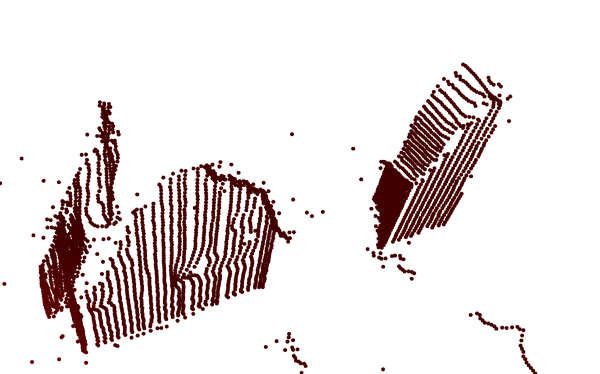}&\includegraphics[width=0.18\textwidth]{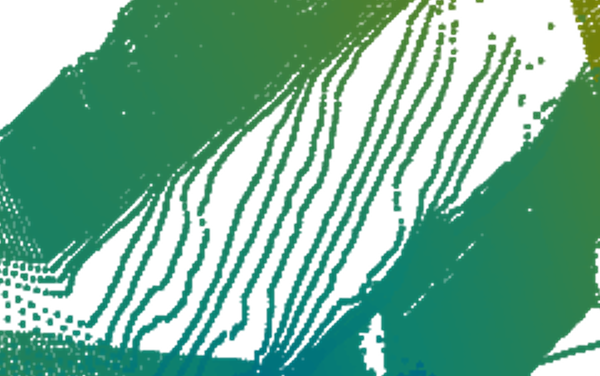}\\
    \rotatebox{90}{\fontsize{7}{10}\selectfont LiDA-SR~\cite{shan2020simulation}}&\includegraphics[width=0.18\textwidth]{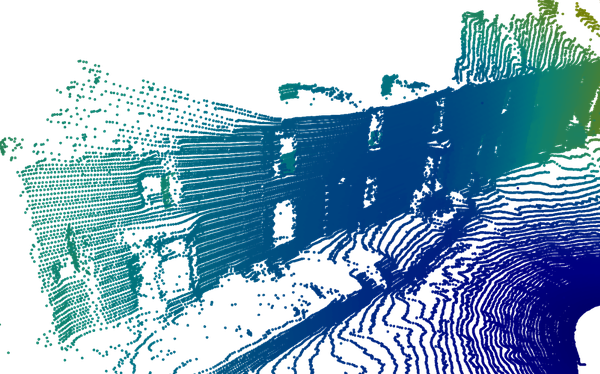}&\includegraphics[width=0.18\textwidth]{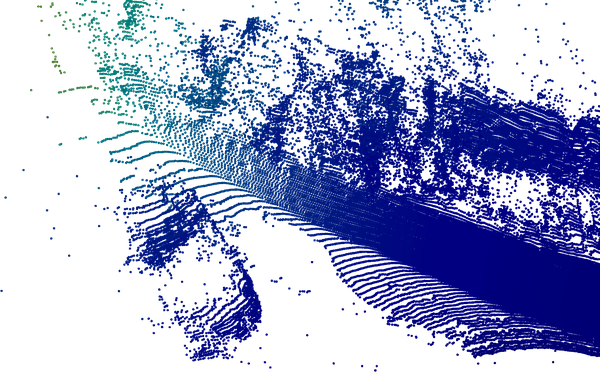}&\includegraphics[width=0.18\textwidth]{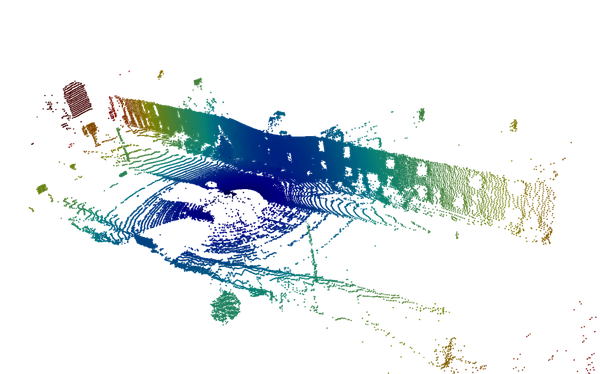}&\includegraphics[width=0.18\textwidth]{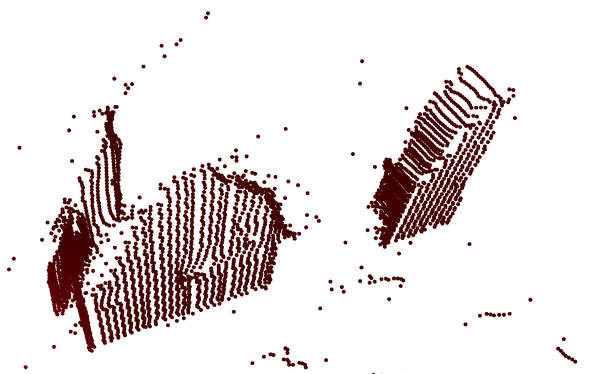}&\includegraphics[width=0.18\textwidth]{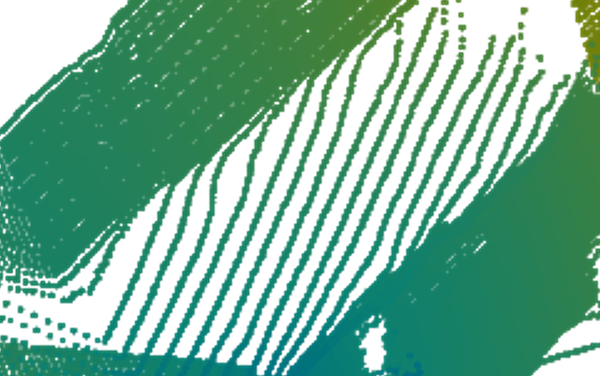}\\
    \rotatebox{90}{\fontsize{7}{10}\selectfont ILN~\cite{kwon2022implicit}}&\includegraphics[width=0.18\textwidth]{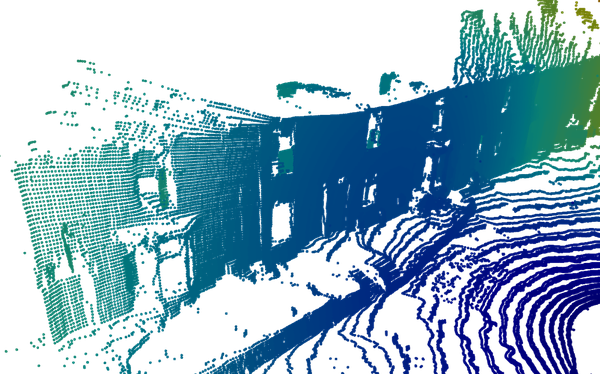}&\includegraphics[width=0.18\textwidth]{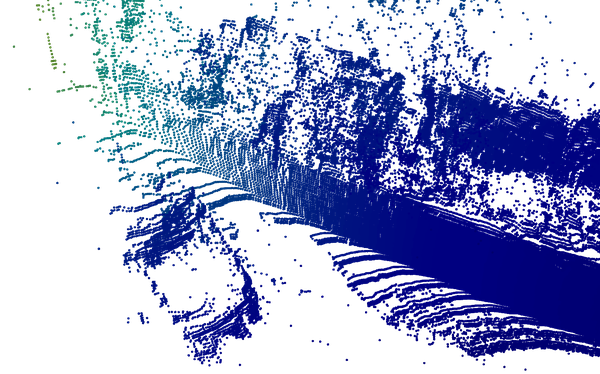}&\includegraphics[width=0.18\textwidth]{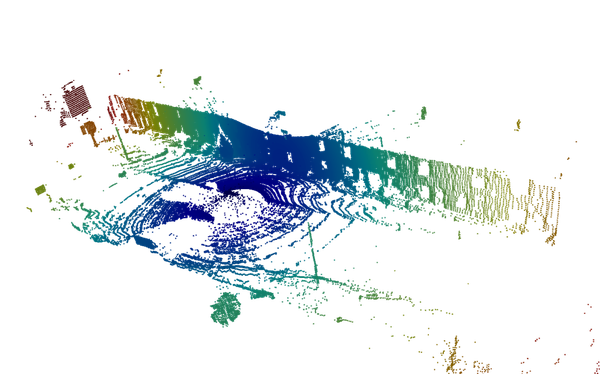}&\includegraphics[width=0.18\textwidth]{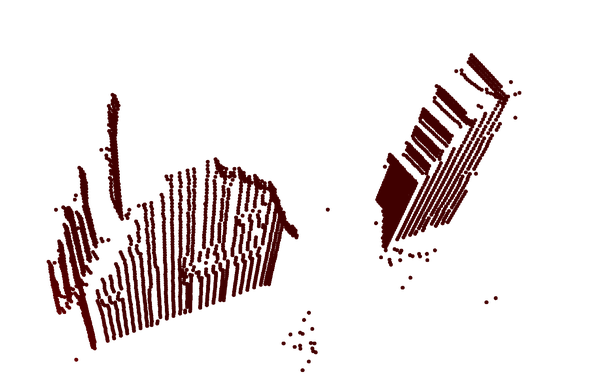}&\includegraphics[width=0.18\textwidth]{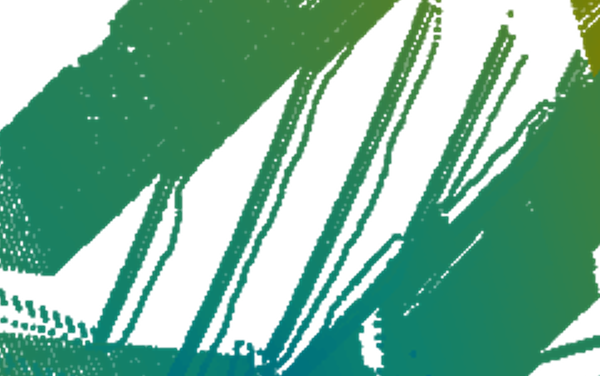}\\
    \rotatebox{90}{\fontsize{7}{10}\selectfont \coolname{}(Ours)}&
    \includegraphics[width=0.18\textwidth]{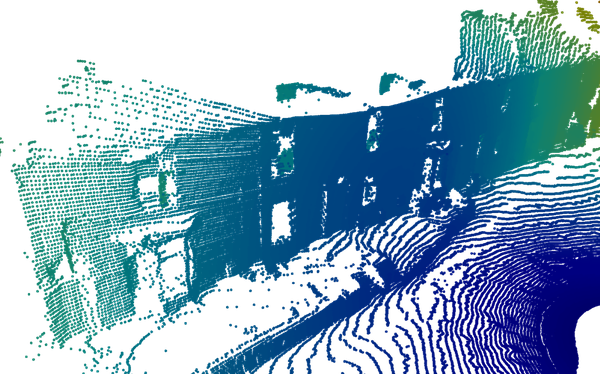}&\includegraphics[width=0.18\textwidth]{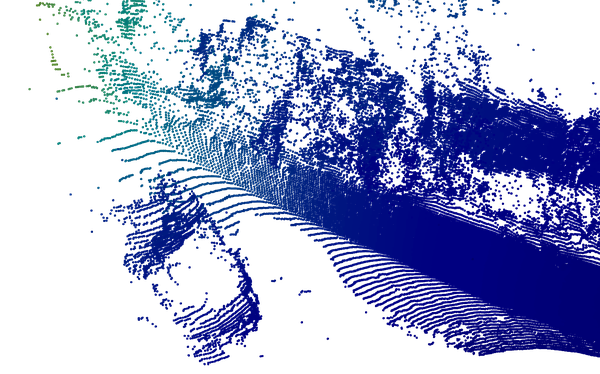}&\includegraphics[width=0.18\textwidth]{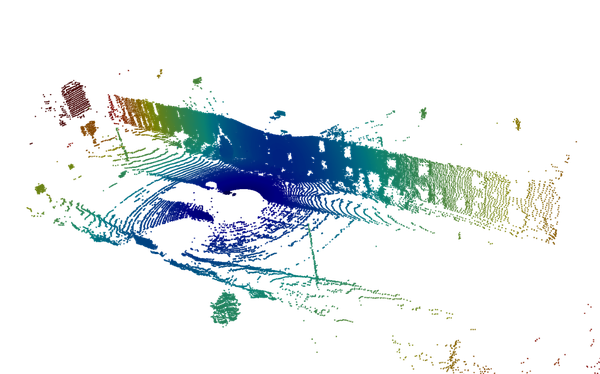}&\includegraphics[width=0.18\textwidth]{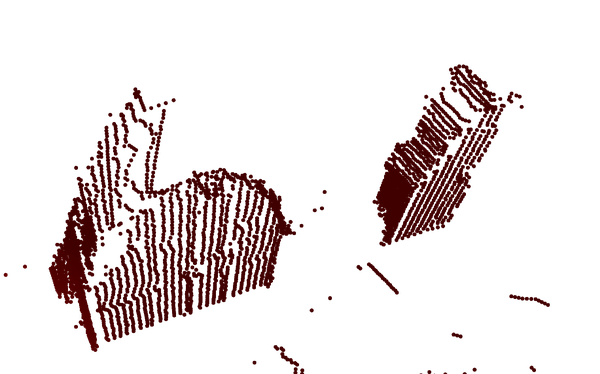}&\includegraphics[width=0.18\textwidth]{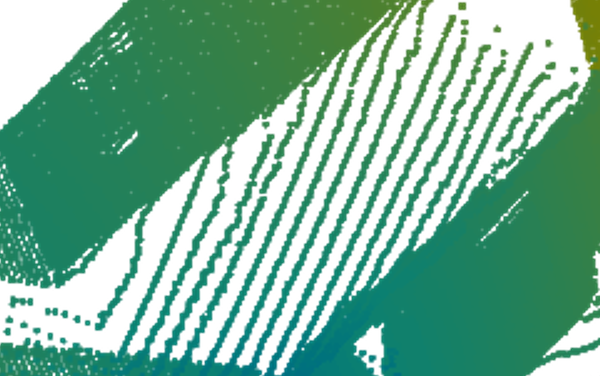}\\
    \rotatebox{90}{\fontsize{7}{10}\selectfont GT}&\includegraphics[width=0.18\textwidth]{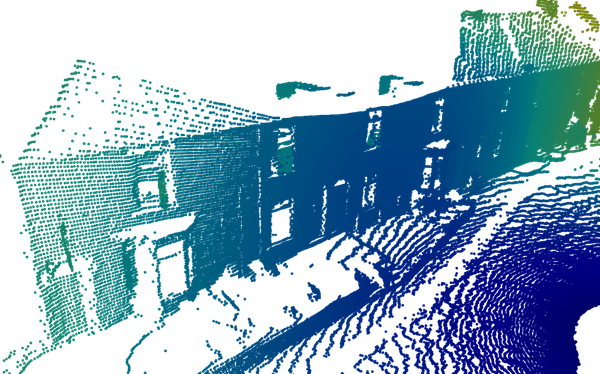}&\includegraphics[width=0.18\textwidth]{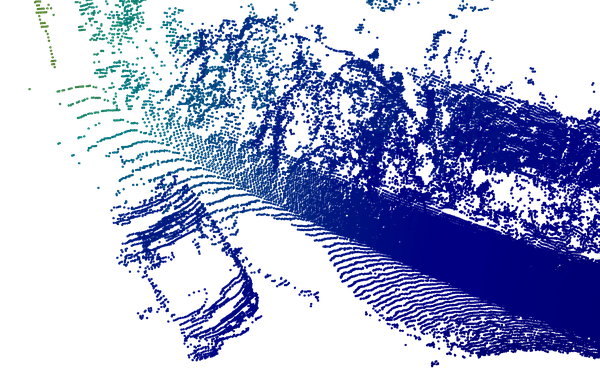}&\includegraphics[width=0.18\textwidth]{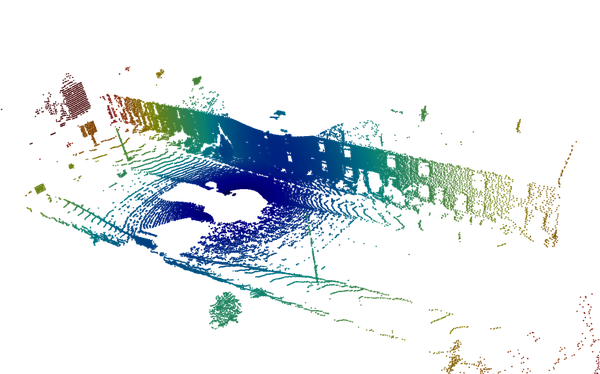}&\includegraphics[width=0.18\textwidth]{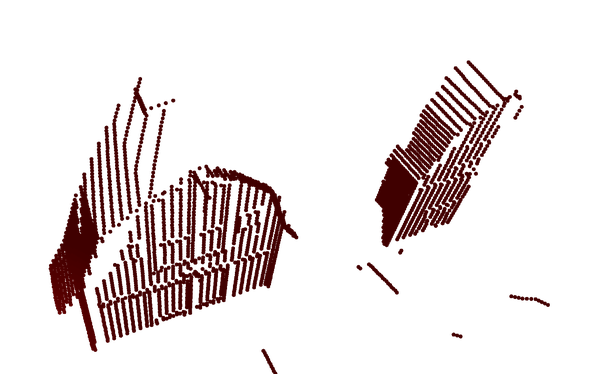}&\includegraphics[width=0.18\textwidth]{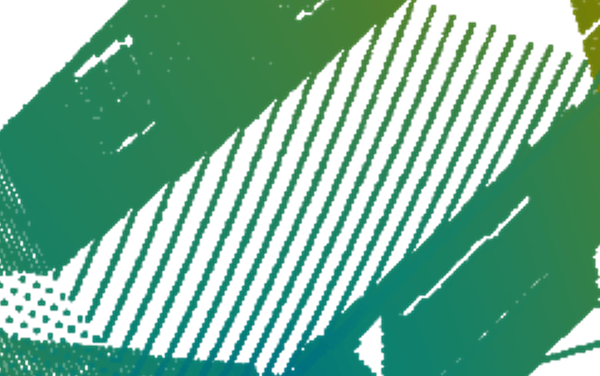}\\
    \quad & \hspace{13mm} F & \hspace{13mm} G & \hspace{13mm} H& \hspace{13mm} I& \hspace{13mm} J\\
    \end{tabular*}
    \caption{Qualitative Results on DurLAR~\cite{li21durlar} (\textbf{F-H}) and CARLA~\cite{kwon2022implicit}(\textbf{I-J}). Our approach can outperform state-of-the-art methods in upsampling scene-related contexts with complex and simple geometry and under both noisy and noiseless circumstances.}
    \label{fig:vis_carla_durlar}
\end{figure*}

\fi

\end{document}